\pdfoutput=1
\documentclass[11pt]{article}

\usepackage[final]{acl}

\usepackage{times}
\usepackage{latexsym}

\usepackage[T1]{fontenc}
\usepackage[utf8]{inputenc}

\usepackage{microtype}

\usepackage{inconsolata}

\usepackage{graphicx}

\usepackage{amsmath}
\usepackage{amssymb}
\usepackage{booktabs}
\usepackage{multirow}
\usepackage{makecell}
\usepackage{subcaption}
\usepackage{enumitem}
\usepackage{siunitx}
\usepackage{tabularx}
\usepackage{listings}

\title{Probing the Geometry of Truth: Consistency and Generalization of Truth Directions in LLMs Across Logical Transformations and Question Answering Tasks}

\author{
~~Yuntai Bao$^{1}$
~~Xuhong Zhang$^{1}$\footnotemark[1]
~~Tianyu Du$^{1}$
~~Xinkui Zhao$^{1}$
~~Zhengwen Feng$^{1}$\\
{\bf ~~Hao Peng$^{2}$}
{\bf ~~Jianwei Yin$^{1}$} \\
    $^{1}$Zhejiang University \\
    $^{2}$Zhejiang Normal University \\
    \texttt{\{yuntaibao, zhangxuhong, zjradty, zhaoxinkui, fengzhengwen\}@zju.edu.cn,} \\
    \texttt{hpeng@zjnu.edu.cn},
    \texttt{zjuyjw@cs.zju.edu.cn}
}

\begin{document}
\maketitle

\renewcommand{\thefootnote}{\fnsymbol{footnote}}
\footnotetext[1]{Corresponding author.}
\renewcommand{\thefootnote}{\arabic{footnote}}

\begin{abstract}
Large language models (LLMs) are trained on extensive datasets that encapsulate substantial world knowledge. However, their outputs often include confidently stated inaccuracies. Earlier works suggest that LLMs encode truthfulness as a distinct linear feature, termed the ``truth direction'', which can classify truthfulness reliably. We address several open questions about the truth direction: (i) whether LLMs universally exhibit consistent truth directions; (ii) whether sophisticated probing techniques are necessary to identify truth directions; and (iii) how the truth direction generalizes across diverse contexts.
Our findings reveal that not all LLMs exhibit consistent truth directions, with stronger representations observed in more capable models, particularly in the context of logical negation.
Additionally, we demonstrate that truthfulness probes trained on declarative atomic statements can generalize effectively to logical transformations, question-answering tasks, in-context learning, and external knowledge sources.
Finally, we explore the practical application of truthfulness probes in selective question-answering, illustrating their potential to improve user trust in LLM outputs.
These results advance our understanding of truth directions and provide new insights into the internal representations of LLM beliefs.\footnote{Our code is public at \url{https://github.com/colored-dye/truthfulness_probe_generalization}}
\end{abstract}

\section{Introduction}
Large language models (LLMs) possess extensive knowledge, as they are trained on immense corpora that encompass a significant portion of world knowledge. However their outputs are not always reliable and are prone to confidently presenting falsehoods \citep{bender2021dangers, evans2021truthful, lin2022truthfulqa, liu2023cognitive}. This unreliability raises critical concerns about the use of LLMs in applications where accuracy is paramount.
A growing body of work \citep{burns2022discovering, azaria2023internal, marks2023geometry, mallen2023eliciting, burger2024truth} aims to elicit accurate information from LLMs despite untruthful outputs. These studies use lightweight classifiers, often referred to as \textit{probes}, to analyze patterns in the model's internal representation that reliably indicate truthfulness. Specifically, given a model and a piece of text, an ideal truthfulness probe is able to tell whether the model believes the text conveys truthful content. By achieving empirical success with linear probes, these works generally believe that truthfulness is internally represented as a salient linear feature and manifests as a ``truth direction''.

The goal of this work is to conduct a more in-depth study of the truth direction as an inherent property of LLMs. Although previous works undoubtedly help us understand truth directions, they fail to answer the following questions: (\textbf{RQ1}) Do LLMs universally represent truthfulness as a linear feature? (\textbf{RQ2}) Are simple probing techniques sufficiently expressive to identify truth directions? (\textbf{RQ3}) If and when a ``truth direction'' exists, in what ways does it generalize? We challenge conclusions from prior works based on empirical evidence, provide preliminary answers to the questions above and present novel observations.

In response to \textbf{RQ1}, we find that not all LLMs exhibit a consistent ``truth direction'', and that this property is closely related to a model's capability. While prior works often assume the universal existence of truth directions, we challenge this assumption. Our evidence suggests that truthfulness is more consistently represented across logical negations in more capable LLMs. Based on this finding, we question the conclusion of \citet{levinstein2024still}, which attributes the generalization failure to limitations in previous probing techniques. Instead, we argue that the inconsistency lies within the LLM itself and that in answer to \textbf{RQ2}, simple supervised probes are sufficiently expressive to identify the truth direction when it is distinctly represented within the model.

In addressing \textbf{RQ3}, we aim to explore the generalization capability of the truth direction, as it illuminates whether an LLM consistently represents truthfulness across different knowledge domains, logical transformations, syntax forms and grounding knowledge source. Previous studies on truth directions have extensively explored the former two aspects of generalization. Regarding syntax forms, they focus either on declarative statements or on $(Q,A)$ pairs; we bridge this gap by testing whether truthfulness probes trained on simple statements generalize to question answering (QA) tasks. Additionally, we examine several variations of QA, including zero- and few-shot QA, with and without provided answer options, and grounded either in parametric knowledge or in question contexts. Our experimental results show that truthfulness probes demonstrate a high degree of generalization.

Furthermore, based on our observation that truthfulness probes are calibrated on certain QA tasks, we introduce their use in selective QA. In this application, we select the subset of answers evaluated as correct by the truthfulness probe from those generated by the LLM. Through this demonstration, we aim to show how truthfulness probes can enhance user trust in real-world LLM-based applications.

Our contributions are as follows:
\begin{enumerate}[topsep=0pt,itemsep=-1ex,partopsep=1ex,parsep=1ex]
    \item We summarize truthfulness probes from prior works, introduce a new instantiation, and conduct extensive experiments.
    \item We study truth directions following three research questions. In addressing \textbf{RQ1}, we explore whether the truth direction is common among LLMs. For \textbf{RQ2}, we test if sophisticated probing techniques are required to identify truth directions. In answer to \textbf{RQ3}, we assess the generalization capabilities of truth directions.
    \item We demonstrate a practical application of truthfulness probes for selective question answering, improving generation quality by filtering out unreliable answers.
\end{enumerate}

\section{Related Work}
\paragraph{Eliciting latent knowledge (ELK).} The field of scalable oversight \citep{christianoeliciting} seeks to address the information asymmetry between superhuman AI systems and human evaluators. It is assumed that although the AI possesses significant knowledge, its behavior is untrustworthy because it is not trained with an objective that explicitly incentivizes outputs to align with the truth \citep{mallen2023eliciting}. ELK is an approach within scalable oversight that aims to identify patterns in an AI's activations that correspond to the truth. The primary challenge lies in identifying patterns that generalize reliably to questions where human evaluators are unable to verify the answers \citep{mallen2023eliciting}. Our work demonstrates that probing techniques can achieve reasonable generalization with limited supervision, suggesting that probes may offer a promising approach for ELK.

\paragraph{Probing for truthfulness.}
Several studies have used probing techniques to uncover truthfulness by examining an LLM's internal states, independent of its inputs or outputs~\citep{lee2023linguistic,joshi2024personas}. Regarding the \textit{geometry of the representation} of truthfulness, most studies agree that this representation is likely linear, as demonstrated by the use of linear probes in studies such as CCS \citep{burns2022discovering}, mass-mean \citep{marks2023geometry,li2023inference}, TTPD \citep{burger2024truth} and the commonly used baseline: logistic regression. Notably, CCS is an unsupervised approach and targets yes/no \textit{question-answering tasks}, while others are supervised and target \textit{factual statements}.
In contrast, some works are \textit{geometry-agnostic}. \citet{azaria2023internal} propose SAPLMA which is based on MLP architecture, while \citet{he2024llm} introduce the LLM Factoscope, which leverages a Convolutional Neural Network architecture.

The aforementioned studies examine truthfulness grounded in a model's \textit{parametric knowledge}, whereas \citet{sky2024androids} detect hallucination with probes in the setting of in-context generation, where \textit{knowledge is grounded in the context}.

\begin{figure*}
    \centering
    \includegraphics[width=.9\linewidth]{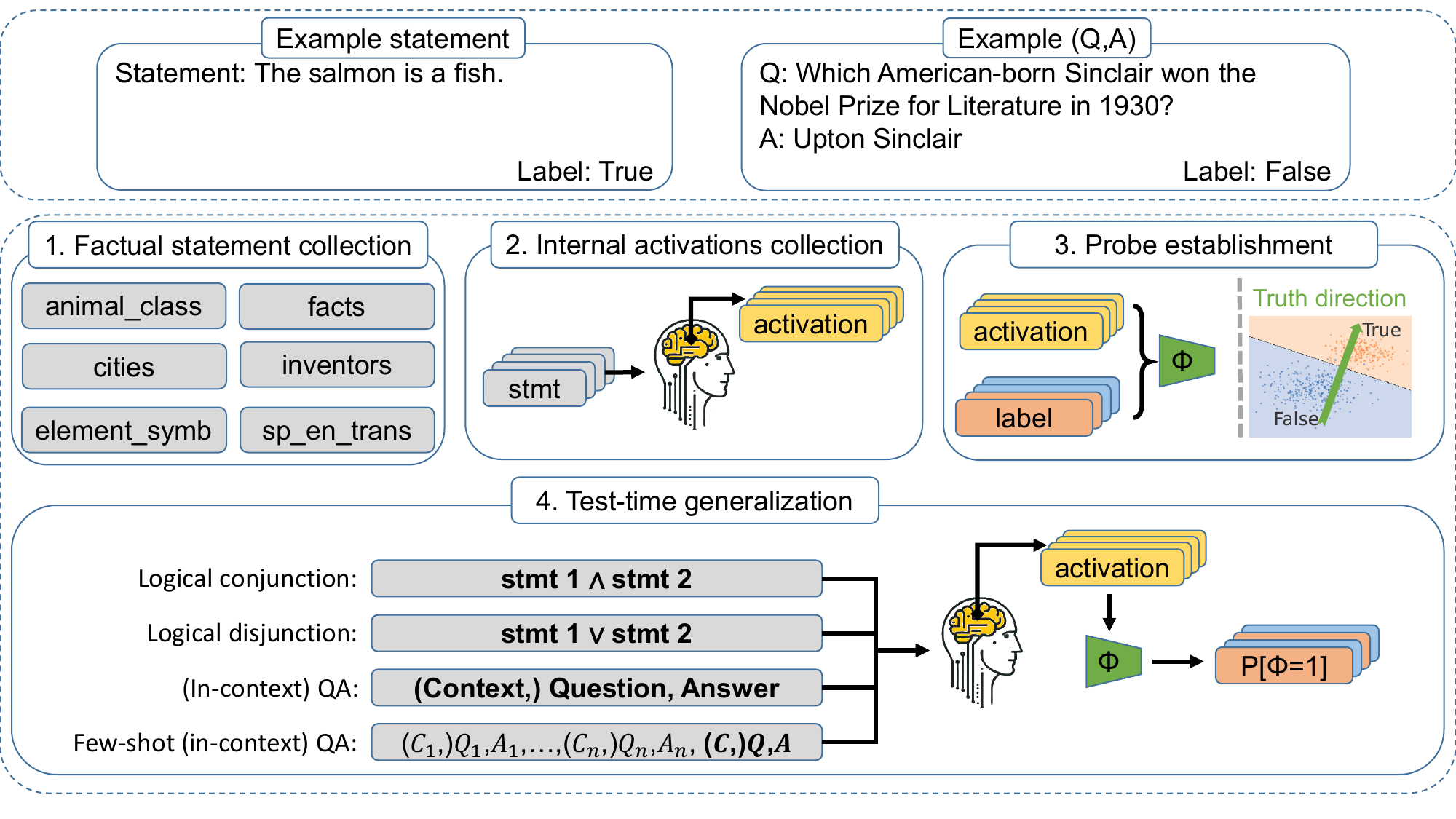}
    \caption{Illustration of truthfulness probes. A truthfulness probe is established using the LLM's internal states when processing labeled statements. The probe is then able to tell if the LLM believes an unseen statement or a given response to a question is true or false leveraging only the LLM's internal states.}
    \label{fig:workflow}
\end{figure*}

\section{Summary of Probes and Data}
In this section, we formally define the task of probing binary features from a model's internal representations and describe the specific probe architectures used in this study. Furthermore, we detail the labeled datasets used to train and evaluate the truthfulness probes.

\subsection{Formulation of Binary Probes}
Our target model is the transformer language model \citep{vaswani2017attention}, which processes token sequences through a series of layers. A sequence of $n$ input tokens, $t=(t_1,t_2,...,t_n)$, is first converted into embeddings, $\boldsymbol{h}^{(0)}=(\boldsymbol{h}^{(0)}_1, \boldsymbol{h}^{(0)}_2,...,\boldsymbol{h}^{(0)}_n)$, by the initial embedding layer. The embeddings are then passed through $L$ layers, where each layer generates representations based on the preceding layer's output. The representation of a single token is a vector: $\boldsymbol{h}^{(j)}_i \in \mathbb{R}^d$ and $(0 \leq j \leq L, 1 \leq i \leq n)$. Finally, the LLM produces predictions using $\boldsymbol{h}^{(L)}$.

Suppose we have prior knowledge that the model internally represents a binary feature. Our goal is to establish a probe $\Phi$, that uses only the model's representations to classify the target attribute. The probe outputs either binary labels $-1/1$, or probabilistic predictions such as $\text{P}[\Phi=1]$.

For autoregressive models, which are the primary focus of this paper, we utilize the representation at the final token position of the $l$-th layer, $\boldsymbol{h}^{(l)}_{-1}$. This approach aligns with prior work \citep{burns2022discovering,azaria2023internal,marks2023geometry}, where the final token position attends to all previous tokens due to the causal attention mechanism. We also assume we have a labeled dataset, $\mathcal{D}=\{(x_i,y_i)\}_{i=1}^{M}$, where $x_i$ represents a token sequence and $y_i$ is the label for the target attribute. Processing these sequences through the model yields $\mathcal{D}_\text{rep}=\{((\boldsymbol{h}^{(l)}_{-1}) _i,y_i)\}_{i=1}^{M}$, which we use to build our classifier.

To maximize classification accuracy, we define a cost function to quantify the classification error of the probe, $J(\Phi,\boldsymbol{h},y)$. The mechanistic objective is to minimize the expected classification error over the data distribution: $\underset{\Phi}{\arg \min} \frac{1}{M} \sum_{i=1}^{M} J(\Phi, \boldsymbol{h}_i,y_j)$.

\subsection{Instantiations of Binary Probes} \label{sec:probe_instantiation}
We classify the probes into two categories: geometry-oriented and statistics-based. Geometry-oriented probes leverage knowledge of the geometric structure of the representation. Under the ``truth direction'' hypothesis, true/false representations can be separated by a hyperplane, and the normal vector of this hyperplace corresponds to the ``truth direction''. In contrast, statistics-based probes are geometry-agnostic and aim to maximize the probability of observing the correct labels given the input data. Our implementations are based on the \verb|scikit-learn| \citep{scikit-learn} library.

\paragraph{Geometry-oriented Probes.}
For geometry-oriented probes, we introduce two instantiations: linear support vector machine (\textbf{SVM}) \citep{cortes1995support} and mass-mean (\textbf{MM}) \citep{marks2023geometry} instantiation. The rationale for selecting linear SVM is its ability to maximize the margin, which aligns with the goal of identifying a separating hyperplane. As SVM does not directly provide probability predictions, we fit a post-hoc probability distribution using Platt scaling through cross-validation on the training data \citep{platt1999probabilistic}.

\paragraph{Statistics-based Probes.}
For statistics-based probes, we present logistic regression (\textbf{LR}) and multi-layer perceptron (\textbf{MLP}). LR is commonly used as a baseline, while MLP is termed SAPLMA by \citet{azaria2023internal}.

\subsection{Data for Probing Truthfulness} \label{sec:data}
The binary probes introduced above are applied to the truthfulness classification task, assuming the availability of truthfulness-specific data. We use the factual statements curated by \citet{burger2024truth}, drawing from datasets by \citet{azaria2023internal} and \citet{marks2023geometry}. These datasets cover a variety of topics, including \verb|animal_class|, \verb|cities|, \verb|element_symb|, \verb|facts|, \verb|inventors|, \verb|sp_en_trans|, as well as variations incorporating logical negations, conjunctions and disjunctions. Each statement is labeled as ``true'' or ``false'', indicating its factuality. Statements can be atomic or compound. Atomic statements make individual claims, either affirmative or negative. Negative statements correspond to their affirmative counterparts, with syntax-level negation and inverted labels. Compound statements are created by logically combining atomic statements of the same topic through conjunction or disjunction.

\section{Experiments} \label{sec:experiments}
\subsection{Preliminary Experiment: Layer Selection} \label{sec:layer_selection}
Identifying the optimal layer for detecting truthfulness is crucial for probe performance. \citet{marks2023geometry} observed that the truth direction ``emerges rapidly in early-middle layers''; however, this observation does not indicate which specific layer provides the most effective representations. To address this, we adopt the technique used by \citet{burger2024truth} and \citet{macdiarmid2024sleeperagentprobes}, which evaluates the difficulty of separating true/false statements across layers by analyzing variance. The ideal layer maximizes the separation between true and false representations, quantified by the ``between-class variance'', relative to the internal variance within each class, referred to as ``within-class variance''. By plotting the ratio of between-class to within-class variance across decoder layers for a range of topic-specific datasets, we identify the optimal layer as the one with the highest ratio.

We present the ratio of between-class variance to within-class variance across layers for Llama-2-7B \citep{touvron2023llama} and Llama-3.1-8B \citep{dubey2024llama} in Figure \ref{fig:separation_across_layers}, with results for additional models in the Appendix. Each curve represents statements involving affirmations, negations, conjunctions and disjunctions of the same topic. For Llama-3.1-8B, the 12th layer (zero-indexed) emerges as the optimal layer. In contrast, for Llama-2-7B, a peak occurs only for the \verb|sp_en_trans| topic, with minimal separation observed for other topics. This suggests that, while Llama-2-7B may internally represent truthfulness as a feature, it does so in a domain-specific manner, with limited consistency across knowledge domains. Additionally, this feature appears to lack salience in the early-middle layers.

\begin{figure}
    \centering
    \begin{subfigure}{1.\linewidth}
        \centering
        \includegraphics[width=1.\linewidth]{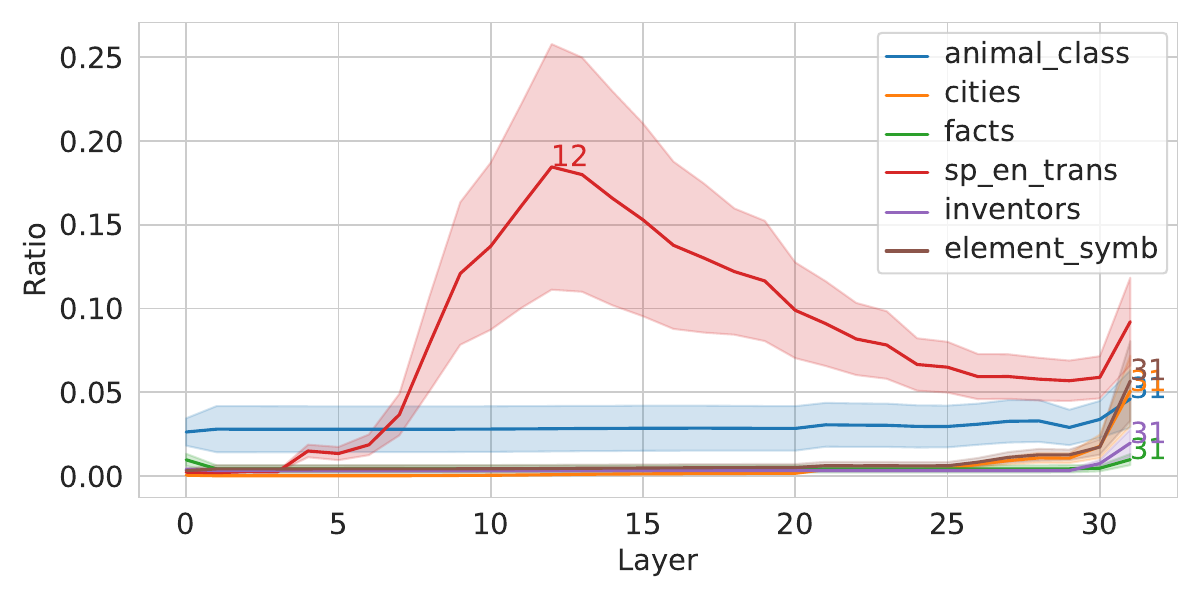}
        \caption{Llama-2-7B}
    \end{subfigure}
    \begin{subfigure}{1.\linewidth}
        \centering
        \includegraphics[width=1.\linewidth]{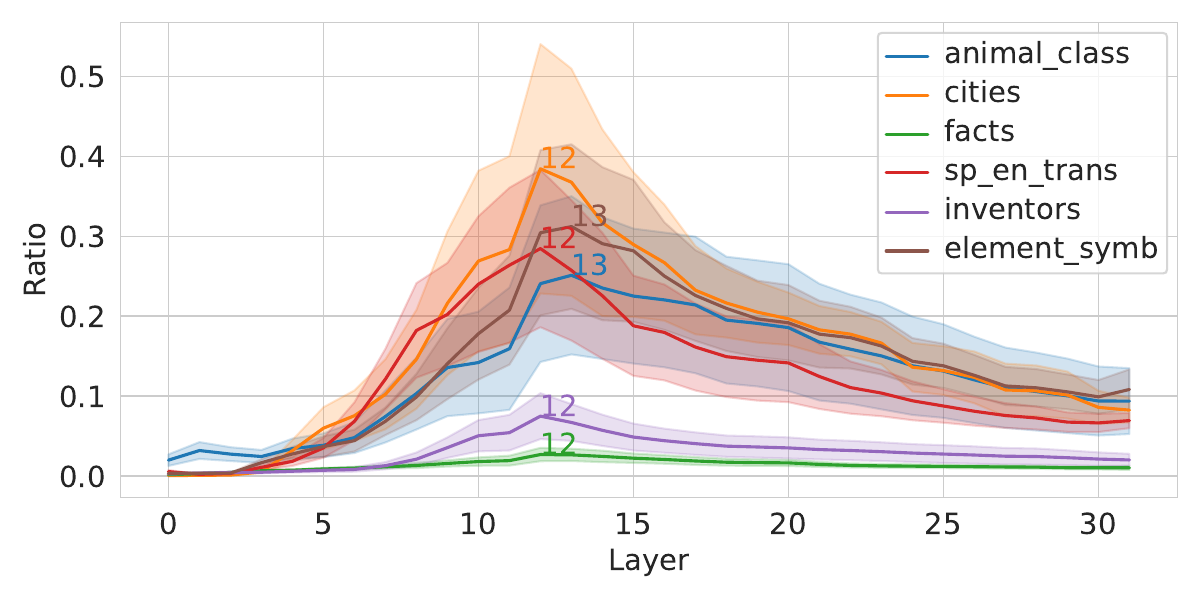}
        \caption{Llama-3.1-8B}
    \end{subfigure}
    \caption{Ratio of between-class variance to within-class variance across layers. The layer indices (starting from 0) for the greatest ratios are annotated at the summit of each curve. The solid curves are mean values, and the surrounding shades denote standard error.}
    \label{fig:separation_across_layers}
\end{figure}

\begin{figure*}[htbp]
    \centering
    \includegraphics[width=1.\linewidth]{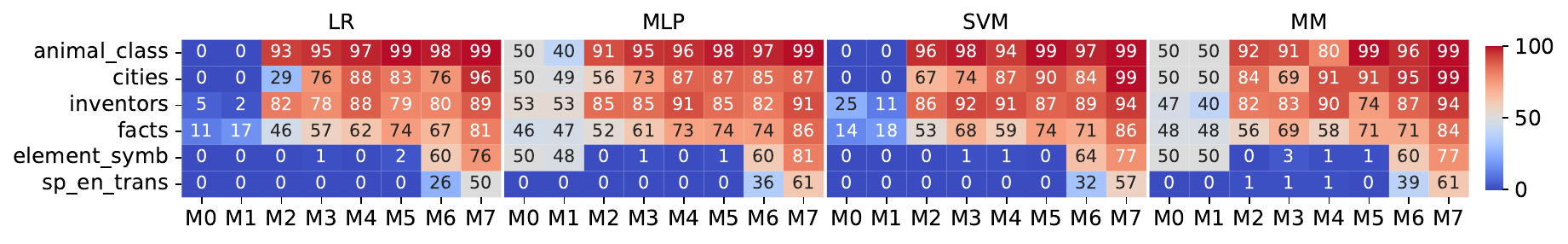}
    \caption{AUROC (in percentage) of probes trained on affirmative statements and tested on negative ones. $\text{AUROC}>0.5$ indicates success of generalization. M0-M7 refer to models from Llama-2-7B to Llama-3.1-70B-Instruct.}
    \label{fig:neg_generalization}
\end{figure*}

\subsection{Probing a Randomized Model}
This section investigates whether the truth direction is an inherent structure within a pretrained LLM or an artifact constructed by the truthfulness probe. To address this, we randomly initialize the weights of Llama-3.1-8B, extract activations from its 12th layer using the \verb|animal_class| dataset, and train probes on a 70\% split while testing them on the remaining 30\%. Results show that the AUROCs are 0.50, 0.52, 0.58, 0.50 for the LR, MLP, MM and SVM probes, respectively. In contrast, when using the pretrained weights, the AUROCs achieve 1.0 across all probes. These findings demonstrate that the simple probes introduced in Section \ref{sec:probe_instantiation} cannot independently construct a truth direction, confirming that truth direction is a product of the pretraining process.

\subsection{Consistency of Truth Directions} \label{sec:rq1}
\citet{levinstein2024still} claim that probes such as MLP fail to generalize across negation. However, we question this conclusion and hypothesize that the generalization performance of truthfulness probes is influenced more by the targeted LLM than by the probe itself. To test this hypothesis, we examine whether the truth direction identified for affirmative statements is consistent with that identified for negative statements of the same topic.

\subsubsection{Experimental Setup} \label{sec:rq1_experiment_setup}
\paragraph{Data.}
For each of the six topics introduced in Section \ref{sec:data}, we train probes on affirmative statements and test them on corresponding negative ones. For example, we train probes on affirmative statements of the \verb|animal_class| topic, and test them on \verb|neg_animal_class|. Note that the training and test data contain the same set of knowledge, differing only in syntax.
\paragraph{Models.}
We select a series of LLMs with increasing levels of general capability, including both foundational models and instruction-tuned ones (According to the evaluation results on a range of standard benchmarks\footnote{\url{https://github.com/meta-llama/llama-models/blob/main/models}}): Llama-2-7B(-Chat), Llama-2-13B(-Chat), Llama-3.1-8B(-Instruct), Llama-3.1-70B(-Instruct). Their optimal layers are 12(13), 13(13), 12(13), 33(33), respectively.
\paragraph{Methods.} We employ the four probe instantiations introduced in Section \ref{sec:probe_instantiation}.
\paragraph{Metrics.}
The metric used is AUROC (Area Under Receiver Operating Characteristic Curve) which is commonly used to assess classifier performance. A probe is considered to generalize successfully on a topic if its test AUROC exceeds 0.5, indicating performance better than chance. Results are averaged across three trials, with randomness introduced through probe initialization and data splits for cross-validation of Platt scaling, while the training data remains constant.

\subsubsection{Results}
The results are presented in Figure \ref{fig:neg_generalization}. We observe a positive correlation between the performance of truthfulness probes and the general capability of the target models. For Llama-2-7B(-Chat), the probes fail to generalize on all six topics. For Llama-2-13B(-Chat) and Llama-3.1-8B(-Instruct), the probes generalize on four topics; for Llama-3.1-70B they generalize on five topics; and for Llama-3.1-70B-Instruct, they generalize on all six topics.

Regarding whether the truthfulness probe is a faithful reflection of the actual truth direction, we borrow the \textit{Weak-to-Strong Explanation} from \citet{zhou-etal-2024-alignment}: if weak classifiers can successfully distinguish the representations, it indicates that LLMs have implicitly converted inputs to different representations. For the most capable model, Llama-3.1-70B-Instruct, all the simple probes we use are able to generalize across logical negation on all six topics. This suggests that Llama-3.1-70B-Instruct consistently represents truthfulness in its internals for both affirmative and negative statements. Therefore the results suggest a potential correlation between the degree of generalization of its truth direction and the model's capability (e.g., knowledge capacity and natural language understanding ability).
Additionally, the differences in performance between probes become negligible starting with Llama-2-13B-Chat and onward. This indicates that, for more capable LLMs, probe performance is more influenced by the target model itself than by the design of the probes.

\subsection{Binary Logical Transformation} \label{sec:logical_conj_disj}
In Section \ref{sec:rq1}, we tested the ability of truthfulness probes to generalize across logical negation. In this experiment, we extend the analysis to more complex binary logical transformations, specifically logical conjunction and disjunction. This requires the LLM to perform several implicit tasks: identify the truthfulness of both atomic statements, interpret binary logical operators from natural language to abstract concepts, and apply the operators to compute the joint truthfulness.

\subsubsection{Experimental Setup}
\paragraph{Data.} The training data consists of all atomic statements. The test data comprises logical conjunctions and disjunctions of atomic affirmative statements for each topic. The knowledge of the test data is covered by the training data.
\paragraph{Models.} We use Llama-3.1-8B as the primary model for demonstration.
\paragraph{Methods.} We apply the four probe instantiations introduced in Section \ref{sec:probe_instantiation}.
\paragraph{Metrics.} For this experiment, we use AUROC as a measure of probe performance. A probe is considered to generalize successfully on a topic if its test AUROC exceeds 0.5. Results are averaged across three trials, with randomness introduced through probe initialization and training data splits.

\subsubsection{Results}

\begin{figure}
    \begin{subfigure}{0.45\linewidth}
        \centering
        \includegraphics[width=\linewidth]{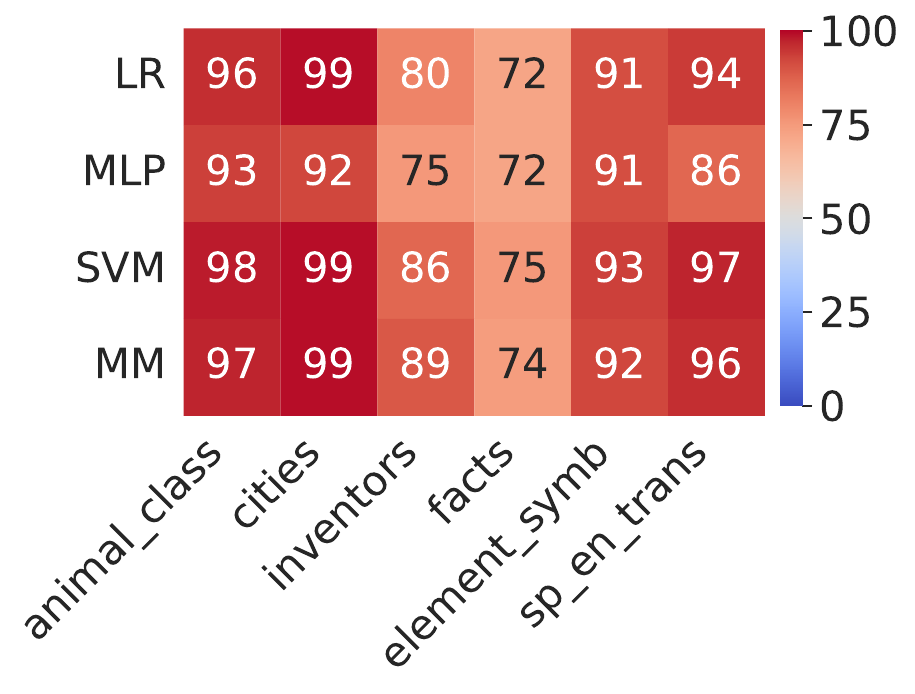}
        \caption{\centering Logical conjunctions.}
    \end{subfigure}
    \begin{subfigure}{0.45\linewidth}
        \centering
        \includegraphics[width=\linewidth]{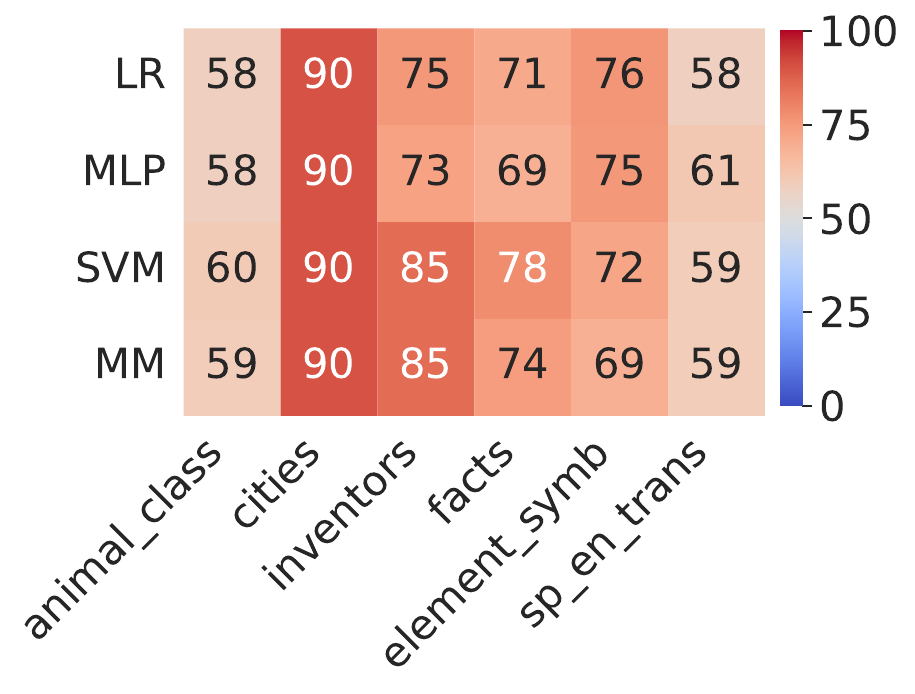}
        \caption{\centering Logical disjunctions.}
    \end{subfigure}
    \caption{AUROC (in percentage) of probes trained on atomic factual statements and tested on logical conjunctions/disjunctions for Llama-3.1-8B. $\text{AUROC}>0.5$ indicates the success of generalization.}
    \label{fig:generalization_to_conj_disj}
\end{figure}

According to Figure \ref{fig:generalization_to_conj_disj}, all probes successfully generalize to both logical conjunctions and disjunctions. However, performance is notably stronger for logical conjunctions compared to disjunctions across \verb|animal_class|, \verb|element_symb|, and \verb|sp_en_trans| topics. This discrepancy may suggest that disjunctions pose a greater challenge for Llama-3.1-8B to interpret truthfulness.

\subsection{Question Answering} \label{sec:generalization_from_stmt_to_qa}
We hypothesize that if truthfulness is consistently represented in an LLM's internal states, this representation should depend solely on the semantics of a sentence, rather than its syntax form. Additionally, question answering is more common than statements in real-world human-AI interactions. Motivated by these considerations, we examine if truthfulness probes, trained on atomic factual statements, can generalize to the QA setting.

We test on a multiple-choice task and a short-form QA task. We also investigate the in-context learning scenario, a popular prompting technique for teaching LLMs new tasks at inference time. While in-context examples can be beneficial, they may include incorrect or misleading examples, which raises questions about how probes handle false examples. Therefore, we pay special attention to the behavior of the probes when incorrect examples are present.

\subsubsection{Experimental Setup}
\paragraph{Data.} The training data consists of all atomic statements. The test data includes MMLU \citep{hendrycks2020measuring} and TriviaQA \citep{joshi2017triviaqa}. For MMLU, we sample 50 questions from the test set for each of the 57 sub-tasks. As it is a multiple-choice dataset, for each question we select the correct answer and an incorrect answer. For TriviaQA, we sample 20 answers per question from the model at unit temperature.
\paragraph{Models.} We use Llama-3.1-8B, as it is a high-capability LLM with $\sim$10B parameters.
\paragraph{Methods.} We apply the four probe instantiations introduced in Section \ref{sec:probe_instantiation}.
\paragraph{Metrics.} In addition to classification accuracy, we evaluate calibration, as it is crucial for assessing the
reliability of predictions. Specifically, we evaluate AUROC, Expected Calibration Error (ECE) and Brier Score (BS). ECE measures calibration, while BS reflects both accuracy and calibration. Lower values are preferred for both ECE and BS. For ECE we use a binned approach with 10 bins, where each bin contains an equal number of samples, and report the mean absolute error between the accuracy and confidence within each bin. The random baseline for BS is 0.25, corresponding to a uniform prediction of 0.5. Results are averaged across three trials, with randomness introduced via probe initialization and training data splits.
\paragraph{Prompt setups.} We test three prompt settings for MMLU: (1) ``zero-shot'': a zero-shot prompt; (2) ``TTTTT'': a five-shot prompt with all correct exemplars; (3) ``TTFFF'': a five-shot prompt where the first two examples are correct and the following three are incorrect. For TriviaQA, we test 5-shot and 20-shot prompts.

Although it is true that the correctness of few-shot examples could be verified in practice, our motivation to study incorrect in-context examples includes: (1) It helps understand how truthfulness probes handle context that contains mixed truthful and untruthful information; (2) It provides insights into how robust truthfulness probes are to potentially conflicting information.

\subsubsection{Results}

\begin{figure}[htbp]
    \centering
    \begin{subfigure}{1.\linewidth}
        \centering
        \includegraphics[width=1.\linewidth]{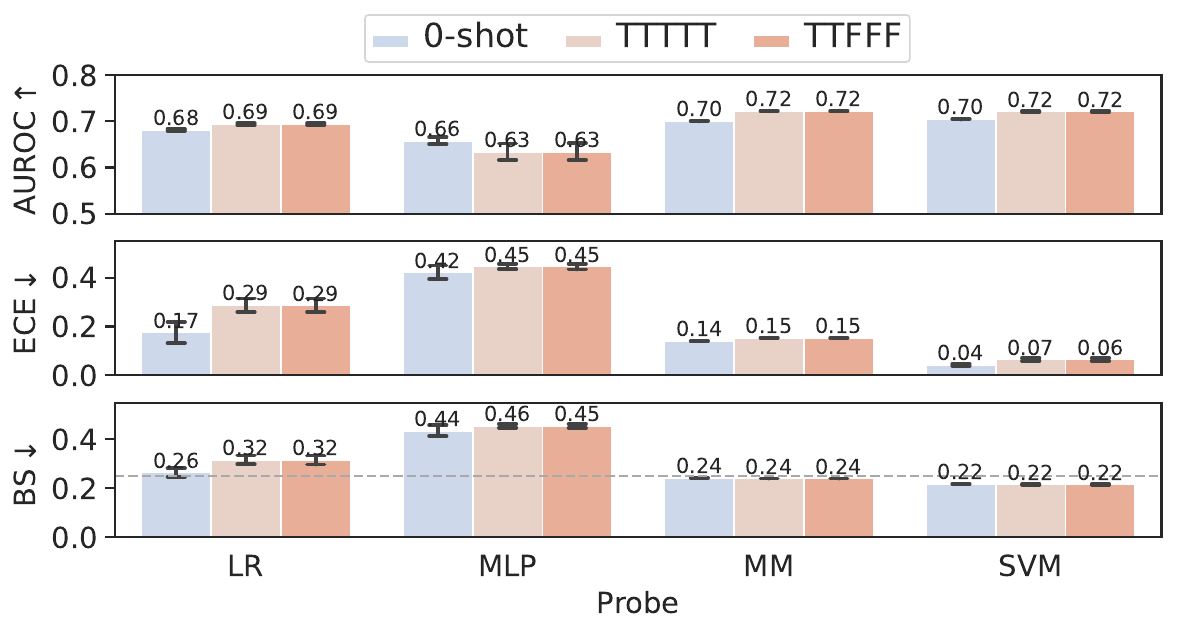}
        \caption{MMLU.}
    \end{subfigure}
    \begin{subfigure}{1.\linewidth}
        \centering
        \includegraphics[width=1.\linewidth]{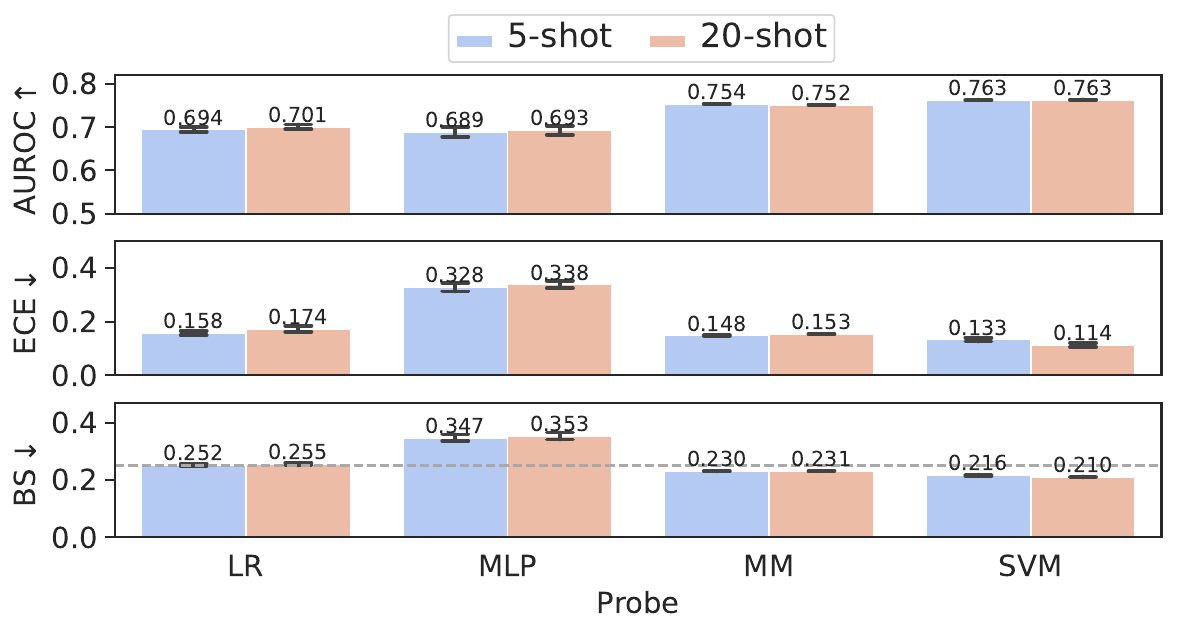}
        \caption{TriviaQA.}
    \end{subfigure}
    \caption{AUROC$\uparrow$/ECE$\downarrow$/BS$\downarrow$ of truthfulness probes for Llama-3.1-8B on MMLU and TriviaQA. The dashed gray line corresponds to random results, and error bars denote standard error.}
    \label{fig:generalization_from_stmt_to_qa_mmlu}
\end{figure}

The results, shown in Figure \ref{fig:generalization_from_stmt_to_qa_mmlu}, reveal that across all probes, accuracy generally improves when few-shot exemplars are provided. This suggests that providing more task-related context in the prompt typically aids the LLM in the implicit truthfulness classification task, and that truthfulness probes not only generalize from factual statements to both multiple-choice QA and short-form QA, but also generalize from fundamental commonsense knowledge to both expert knowledge (MMLU) and trivia knowledge (TriviaQA). Notably, according to the results on MMLU, the effect of imperfect few-shot prompts is near identical to entirely correct few-shot prompts. This indicates that the performance of truthfulness probes is not significantly influenced by the truthfulness of the few-shot examples -- the probes express the truthfulness of the \textit{final} $(Q,A)$ pair, treating prior exemplars as context.
Interestingly, this finding aligns with \citet{halawi2023overthinking}'s observation that early layers of the LLM are insensitive to false in-context demonstrations.

However, few-shot prompting does not always improve calibration, and a positive case is only observed for the SVM probe on the TriviaQA dataset. Among all probes, the SVM probe performs best in terms of both classification accuracy and calibration, likely because of its ability to accurately identify the geometry of the truth direction and the benefits of the additional Platt scaling procedure.

\subsubsection{Additional Experiments}
Beyond the generalization of probes trained on factual statements, we are interested in how these probes differ from in-domain probes, which are trained and tested on data of the same domain.
Therefore we perform follow-up experiments by training and testing probes on MMLU using the Llama-3.1-8B model, using the same data and prompt setup as above.  
Specifically, we train on a random 70\% split and test on the rest 30\%.

\begin{figure}[tbp!]
    \centering
    \includegraphics[width=\linewidth]{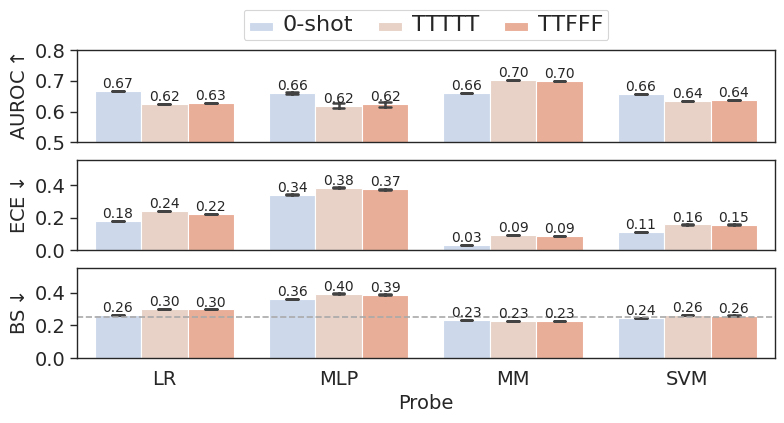}
    \caption{AUROC$\uparrow$/ECE$\downarrow$/BS$\downarrow$ of in-domain probes for Llama-3.1-8B trained and tested on MMLU.}
    \label{fig:mmlu_in_domain}
\end{figure}

We present results averaged over three random trials in Figure \ref{fig:mmlu_in_domain}.
Comparing the results with those of Figure \ref{fig:generalization_from_stmt_to_qa_mmlu}, we observe that in-domain probes generally under-performs truthfulness probes trained on atomic factual statements in terms of AUROC. This finding indicates that truthfulness probes under the current setup perform better than those under the in-domain QA setup.

\subsection{Contextual Knowledge} \label{sec:generalization_from_parametric_knowledge_to_contextual_knowledge}

\begin{figure*}[tb!]
    \centering
    \includegraphics[width=\linewidth]{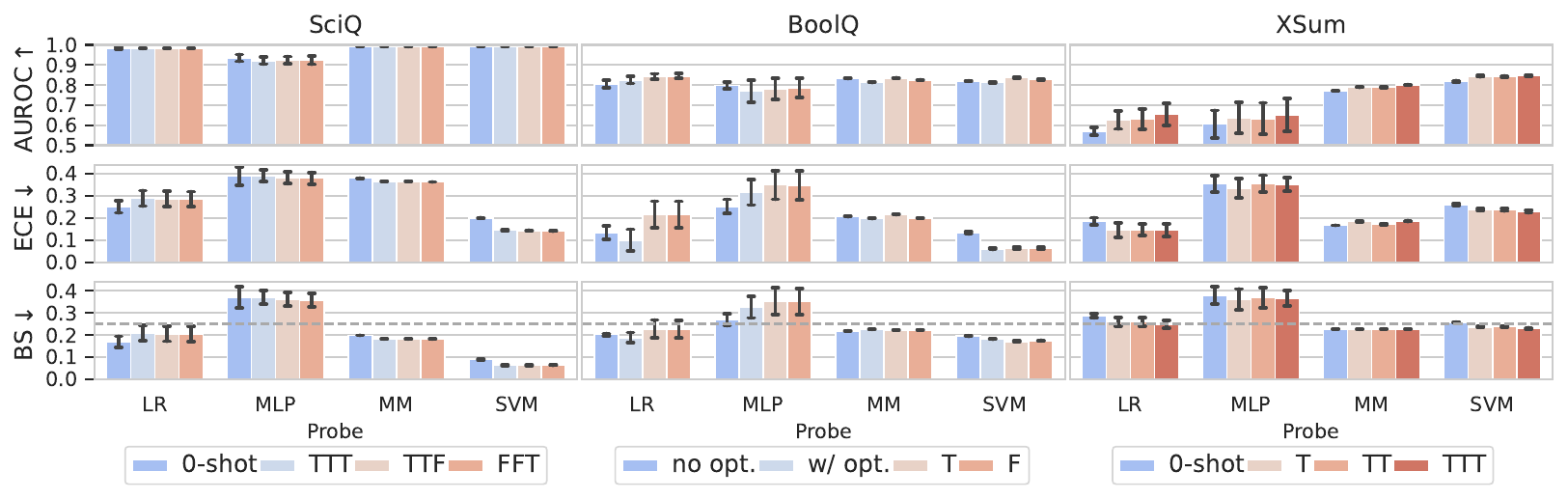}
    \caption{AUROC$\uparrow$/ECE$\downarrow$/BS$\downarrow$ of truthfulness probes for Llama-3.1-8B on tasks where grounding knowledge is provided in the prompt. The dashed gray line corresponds to random results, and error bars denote standard error.}
    \label{fig:generalization_from_paramatric_to_contextual}
\end{figure*}

When training truthfulness probes using factual statements, we are targeting the factual aspect of truthfulness, where the grounding knowledge resides in the LLM's parameters. However, in-context knowledge also plays a vital role in generation.
In this section, we investigate if truthfulness probes can also capture this additional aspect of truthfulness, where the grounding knowledge is provided as contextual information in the prompt.
Notably, context grounding is fundamentally different from factual correctness, since faithfully following false context is acceptable for the former but not for the latter.
We conduct experiments on two tasks: in-context QA and abstractive summarization.

\subsubsection{Experimental Setup}
\paragraph{Data.} The training data consists of all the atomic statements. For the in-context QA task we use the SciQ \citep{welbl2017crowdsourcing} and BoolQ \citep{clark-etal-2019-boolq} datasets. For SciQ, we randomly select 1000 questions, pairing each question with the lettered choice for both the true and a randomly selected false answer. The answers of BoolQ are binary ``yes/no'', therefore we flip the answers to balance true and false $(Q,A)$ pairs. For the abstractive summarization task, we use the XSum dataset \citep{Narayan2018DontGM} and XSum Hallucination Annotations \citep{maynez-etal-2020-faithfulness}. For each article, we pair true summaries from the former dataset with false ones from the latter one.
\paragraph{Models.} We use Llama-3.1-8B.
\paragraph{Methods.} We apply the four probe instantiations introduced in Section \ref{sec:probe_instantiation}.
\paragraph{Metrics.} We report AUROC, ECE and BS. The results are averaged over three trials, with randomness introduced via initialization and data splits.
\paragraph{Prompt setups.} For SciQ dataset which is a multiple-choice task, we implement four settings: (1) ``zero-shot'': zero-shot prompt; (2) ``TTT'': three-shot prompt where all exemplars are correct; (3) ``TTF'': three-shot prompt, where the first two examples are correct and the third is incorrect; (4) ``FFT'': three-shot prompt, where the first two examples are incorrect and the third is incorrect. For BoolQ, we implement four settings: ``no options'', ``with options'', and one-shot ``T''/``F'' (with possible options). For XSum, no options are provided in the prompt as it is not a multiple-choice task. We implement the following prompt configurations: zero-shot, ``T'', ``TT'', and ``TTT''.

\subsubsection{Results}
The results are shown in Figure \ref{fig:generalization_from_paramatric_to_contextual}. Across all datasets and most probes, possible answer options and few-shot exemplars generally improve classification accuracy. This indicates that truthfulness probes generalize from factual statements to both in-context multiple-choice QA tasks and abstractive summarization tasks.

Among the probes, statistics-based probes (LR and MLP) display greater standard error in terms of all three metrics than geometric-oriented ones (MM and SVM), likely due to their optimization instability, and the SVM probe performs best from the perspective of BS.
Additionally, discrepancies are observed for LR probe on the SciQ and BoolQ datasets, MLP probe on the BoolQ dataset, and MM probe on the XSum dataset, where accuracy improves in response to in-context exemplars but calibration worsens. We assume that these discrepancies could be explained with the overconfidence of the probes' predictions.

\subsection{Selective Question Answering}
Based on the findings of Section \ref{sec:generalization_from_stmt_to_qa}, we observe that truthfulness probes trained on atomic statements are capable of generating calibrated probabilistic predictions for QA tasks while achieving reasonable accuracy. Building upon these observations, this section investigates whether truthfulness probes can be leveraged to selectively identify correct answers from a set of candidate responses sampled from an LLM. This setup is inspired by the work of \citet{kadavath2022language}, who demonstrated that an LLM can evaluate the correctness of its own answers through verbal feedback.
While our selective QA experiment shares similarities with ~\citet{kadavath2022language}, our approach differs fundamentally as we leverage internal representations rather than model-generated feedback.

For this experiment, we use the TriviaQA test data from Section \ref{sec:generalization_from_stmt_to_qa}, where we sample 20 answers from the Llama-3.1-8B model using a 20-shot prompt with unit temperature. To perform selective QA, we select the subset of $(Q,A)$ pairs for which the truthfulness probe predicts $\text{P}[\Phi=1]>0.5$ and report the accuracy on this subset. We use the SVM probe, as it performs best in terms of both classification accuracy and calibration.

The aggregated accuracy across all sampled $(Q,A)$ pairs is 55.29\%. Among these, the truthfulness probe classifies 80.26\% as true, and the accuracy of this subset is 64.06\%. This demonstrates that truthfulness probes can be used to filter out false answers sampled from LLMs.

% \subsubsection{Experimental Setup}
% \paragraph{Data.} We use the TriviaQA dataset, as detailed in Section \ref{sec:generalization_from_stmt_to_qa}, where we sample 20 answers from Llama-3.1-8B using a 20-shot prompt with unit temperature.

% \paragraph{Models.} We employ the Llama-3.1-8B model as a representative example, as it provides a robust basis for exploring truthfulness probes in the context of selective question answering.

% \paragraph{Methods.} To perform selective QA, we select a subset of $(Q,A)$ pairs for which the truthfulness probe predicts $\text{P}[\Phi = 1] > 0.5$. We report the accuracy of the probe's classification on this subset. Among the probe instantiations, we choose the SVM probe, as it consistently performs best in terms of both accuracy and calibration.

% \paragraph{Metrics.} We evaluate accuracy on the subset of $(Q,A)$ pairs classified by the probe as true.

% \subsubsection{Results}
% The aggregated accuracy of all sampled $(Q,A)$ pairs is 55.29\%. Among these, the truthfulness probe classifies 80.26\% as true, and the accuracy for this subset is 64.06\%. These results show that truthfulness probes can effectively enhance the accuracy of answers selected from a set of samples generated by an LLM, providing a means for improving the reliability of sampled answers.

\section{Conclusions}
In this work, we provide preliminary evidence supporting our hypothesis that consistent truth directions only emerge in capable LLMs and not in weaker ones, and they could be effectively identified with simple linear probes. We also investigate the generalization properties of truth directions. Empirical results show that truthfulness probes trained only on atomic statements generalize well to logical transformations, (few-shot) question answering and contextual truthfulness. These findings underscore the potential of truthfulness probes to identify truth directions using simple anchor data, thereby facilitating the elicitation of latent knowledge within LLMs.

\section{Limitations}
This study has several limitations that warrant discussion. First, the term "truth" as used in this paper represents an idealized concept, but it may not be what the truthfulness probes actually measure. Drawing on \citet{kadavath2022language} and \citet{marks2023geometry}, the pretraining of language models largely involves imitating human-generated text. Consequently, truthfulness probes are likely capturing an overlap between widely accepted human beliefs and factual, objective truths about the physical world. This raises interesting questions about how such probes might perform in the context of scalable oversight, particularly with hypothetical AI systems that surpass human intelligence.

Second, our investigation into the generalization of truthfulness probes is limited to short-form QA. Extending this analysis to more complex scenarios, such as long-form QA or instruction-following tasks, may yield novel insights and uncover more practical applications of truthfulness probes.

Third, the causality of truthfulness probes is unclear. Our approach relies on classic classification techniques, consistent with prior work on probing truthfulness. Meanwhile, we do not discuss the causal effects of truth directions, i.e., whether LLMs utilize the implicit truthfulness classification results for predictions. \citet{marks2023geometry} conducted causal intervention experiments using the mass-mean probe and showed that mass-mean directions are highly causal. However, as highlighted by \citet{kumar2022probing}, probes often capture spuriously-correlated features rather than exclusively isolating the target feature. Future research on the causal implications of truth directions could provide valuable insights into guiding LLMs to produce more truthful responses, as in \citet{li2023inference}.

Finally, our experiments are constrained by computational resources, with the largest model evaluated being Llama-3.1-70B. As a result, our hypothesis that highly capable LLMs will eventually establish a consistent internal concept of truthfulness remains untested on more advanced models such as GPT-4 \citep{achiam2023gpt}.

\section*{Acknowledgment}
This work was partly supported by the National Key Research and Development Program of China under No. 2024YFB3900105, NSFC under No. 62402418, Zhejiang Province's 2025 “Leading Goose + X” Science and Technology Plan under grant No.2025C02034, the Key R\&D Program of Ningbo under No. 2024Z115, and the Open Project of Key Laboratory of General Quality Technology and Application of Intelligent Manufacturing Equipment, Ministry of Industry and Information Technology (HK202403532).

\bibliography{custom}

% \clearpage
% \newpage
\appendix

\section{Truth-related concepts}
In this section, we distinguish the term ``truthfulness'' from several related concepts. Truthfulness refers to the alignment of a statement or $(Q,A)$ with either world knowledge or contextual sources. The former, following \citet{mahaut-etal-2024-factual}, is termed ``\textit{factuality}''. Contextual truthfulness, by contrast, may include fictional information that deviates from real-world facts, such as solving math problems in a hypothetical scenario.

Untruth lies at the negative end of the truthfulness spectrum and differs from \textit{hallucination}, which refers to generations that are nonsensical or unfaithful to the provided source content \citep{ji2023survey}. A key distinction between untruth and hallucination is that truthfulness requires a sentence to be both sensical and unambiguous. Additionally, we differentiate untruth from \textit{lies}. According to \citet{pacchiardicatch}, an answer is considered a lie only if the speaker knows the correct answer. In this view, a lie is a subset of untruths.

% We distinguish truthfulness from related concepts. Truthfulness refers to a statement's alignment with either world knowledge or contextual sources, and the former is termed "factuality" \citep{mahaut-etal-2024-factual}. Hallucination, on the other hand, refers to nonsensical or unfaithful generations \citep{ji2023survey}. A lie, as defined by \citet{pacchiardicatch}, is a subset of untruths where the speaker knows the correct answer.

\section{Explanation on Choice of Probes}
In this work we summarize LR, MLP, SVM and MM instantiations and use them for experiments. We do not use the TTPD probe introduced by the recent work, \citet{burger2024truth}, for two reasons. First, the design of the TTPD probe is based on their finding that the ``affirmative truth direction'' and the ``general truth direction'' are not aligned. However, according to our findings in Section \ref{sec:rq1}, the ``affirmative truth direction'' and the ``general truth direction'' become more consistent as the target model's general capability increases. Therefore, for models with relatively high capability, the TTPD probe is not expected to distinguish itself among other probe instantiations.

\begin{figure}[hbt!]
    \centering
    \includegraphics[width=1.\linewidth]{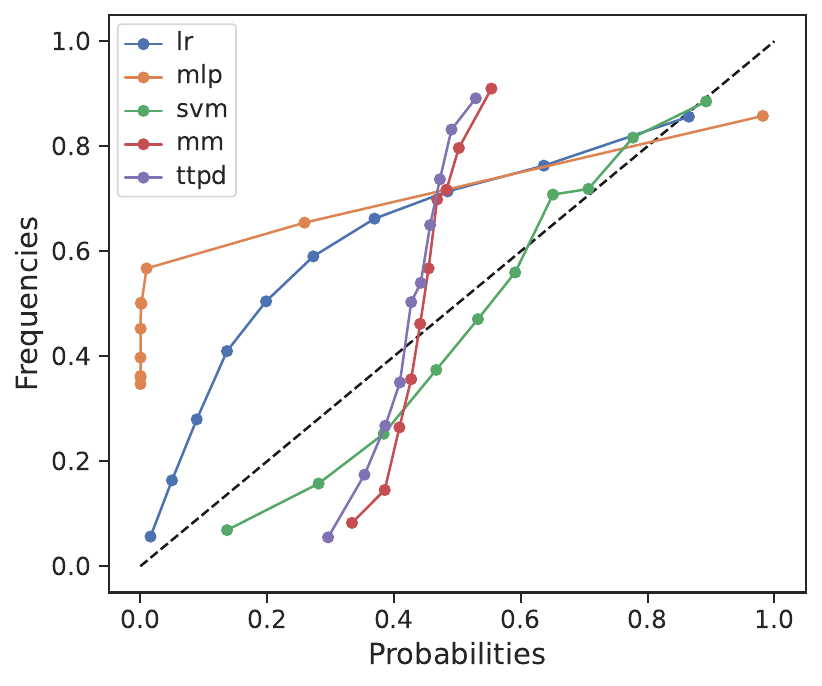}
    \caption{Calibration graph of LR, MLP, SVM, MM and TTPD probes on the BoolQ dataset under the ``with options'' setting. The target LLM is Llama-3.1-8B.}
    \label{fig:calibration_curve_boolq_with_options}
\end{figure}

\begin{figure*}[htbp]
    \centering
    \includegraphics[width=1.\linewidth]{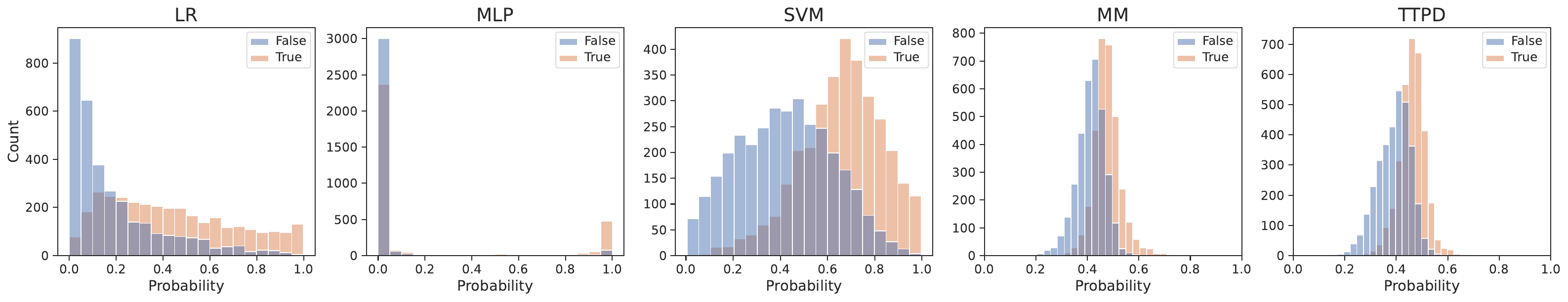}
    \caption{Output distributions of LR, MLP, SVM, MM and TTPD probes on the BoolQ dataset under the ``with options'' setting. The target LLM is Llama-3.1-8B.}
    \label{fig:prob_dist_boolq_with_options}
\end{figure*}

Second, we find empirical evidence that the TTPD probe is not the most effective probe. We replicate the experiment on the BoolQ dataset in Section \ref{sec:generalization_from_parametric_knowledge_to_contextual_knowledge}, and plot the output distribution and calibration curves of all probes targeting Llama-3.1-8B under the ``with options'' setting. The results are shown in Figure \ref{fig:prob_dist_boolq_with_options} and Figure \ref{fig:calibration_curve_boolq_with_options}. The output distribution of the TTPD probe resembles that of the MM probe, so does the calibration curve. This observation is not restricted to this setting and the BoolQ test set but is also noticed in a number of other settings and test sets. These findings suggest that judging by the functional behaviors of the TTPD probe, its performance is not representative enough to be reported.

\section{Details of Factual Datasets}
We mentioned in Section \ref{sec:data} about the factual statements covering six topics which are used for training and testing truthfulness probes. We summarize each topic-specific dataset in Table \ref{tbl:dataset_summary}, according to the data curators \citet{azaria2023internal}, \citet{marks2023geometry} and \citet{burger2024truth}.

Instead of directly using the statements from \citet{burger2024truth}, we perform minor modifications on the \verb|inventors| topic and on logical disjunctions. We notice that the original \verb|inventors| dataset has potential ambiguity due to duplication of name. Thus we specify that the person mentioned in a statement was an inventor, e.g. from ``Thomas Edison lived in the U.S.'' to ``The inventor Thomas Edison lived in the U.S.''. Another tweak is that the original logical disjunctions composed by \citet{burger2024truth} are not consistent with conjunctions, as the subjects are written in full for conjunctions while the subjects are written in pronouns for disjunctions. To align these two logical transformations, we recover the subjects for disjunctions.

\begin{table*}[htbp]
\centering
\begin{tabular}{l l l}
\toprule
{\bf Topic} & {\bf Description} & {\bf Example statement} \\
\midrule
\verb|animal_class| & The class of a specific animal species. & The salmon is a fish.\\ \hline
\verb|cities| & Locations of world cities. & The city of Krasnodar is in Russia. \\ \hline
\verb|element_symb| & \makecell[l]{Chemical elements and\\ their abbreviations.} & Thallium has the symbol Tl. \\ \hline
\verb|facts| & Diverse scientific facts. & \makecell[l]{The Earth's atmosphere protects\\ us from harmful radiation\\ from the sun.}\\ \hline
\verb|inventors| & Home countries of inventors. & \makecell[l]{The inventor Edwin\\ Herbert Hall lived in the U.S.}\\ \hline
\verb|sp_en_trans| & Translations of Spanish words to English. & \makecell[l]{The Spanish word 'con'\\ means 'to speak'.} \\
\bottomrule
\end{tabular}
\caption{Summary of topic-specific factual statement datasets.}
\label{tbl:dataset_summary}
\end{table*}

\section{Probe Implementation Details}
Our implementation of the LR, SVM and MLP probes is based on the \texttt{scikit-learn} \citep{scikit-learn} library. For the LR probe, we employ the L-BFGS optimization algorithm \citep{liu1989limited}. For SVM, we utilize the \verb|NuSVC| implementation. We set $\nu=0.5$, a choice later validated by experiment results. Platt scaling is applied using five-fold cross-validation with the help of the \texttt{scikit-learn} library. For the MLP probe we configure a decreasing sequence of hidden units (512,128,64) with \texttt{tanh} activation, and we use the Adam optimizer \citep{kingma2014adam} to train it till convergence. Finally, we use the MM probe implementation provided by \citet{marks2023geometry}.

When establishing probes on atomic statements, we use a random 70\% split for training and hold out the rest as the development set.

\section{Metric Details}
\subsection{Expected Calibration Error (ECE)}
For ECE, we first sort the probabilities predicted by a probe and split them into $N$ equal-sized bins. In this paper we let $N=10$, which is common in literature evaluating calibration. For each bin, we calculate the mean probability ($x_i$) and the fraction of truthful predictions ($y_i$). ECE is then computed following the formula below:
\begin{equation}
    \text{ECE} = \frac{1}{N} \sum_{i=1}^{N} |y_i - x_i|.
\end{equation}

\subsection{Brier Score (BS)}
The Brier Score measures the difference between the actual correctness and the confidence score through point-wise mean squared error. Its formulation is as follows:
\begin{equation}
    \text{Brier Score} = \frac{1}{N} \sum_{i=1}^{N} (p_i - y_i)^2,
\end{equation}
where $p_i$ is the confidence reported by the probe and $y_i \in \{0,1\}$ is the ground truth label. When a predictor is always making inconfident random predictions, i.e. $p_i=0.5 (i=1,2,...N)$, it results in a chance Brier score of 0.25.

\section{Experiment Details and More Results}
In this section, we elaborate on the detailed setups of the experiments in Section \ref{sec:experiments}. Furthermore, as we only use the Llama family of models in the body of the paper, in this section we also demonstrate results on a model of the Mistral family, Mistral-7B-v0.1 \citep{jiang2023mistral}.

\subsection{Computational and Storage Resources}
All of our experiments are completed on three A6000 (48GB) GPUs. For most LLMs except Llama-3.1-70B(-Instruct), only one GPU will suffice. However, the storage for activations across all datasets and models would take $\sim$1TB disk space. Therefore we recommend modifying the code and only keep the necessary activations on disk. Furthermore, in order to accommodate large LLMs such as Llama-3.1-70B(-Instruct) into our GPUs when gathering hidden activations, we use \verb|float8| quantization with the \verb|optimum-quanto|\footnote{\url{https://github.com/huggingface/optimum-quanto}} library.

\subsection{Selecting Layer}
In Section \ref{sec:layer_selection} we discuss the selection of decoder layer residual stream to extract truth direction from. We present a criteria based on the ratio of between-class variance to within-class variance. However, due to limitation of space we only present results for Llama-2-7B and Llama-3.1-8B in Section \ref{sec:layer_selection}. Here we replicate these approaches on more models.

The data used for plotting is the collection of both affirmative and negative atomic statements covering all the six topics, as well as their logical conjunctions and disjunctions. The curve for each topic consists of the four variations of statements, which results in six curves for each model.

\begin{figure*}
    \centering
    \begin{subfigure}{.48\linewidth}
        \includegraphics[width=\linewidth]{figures/Llama-2-7b-hf_no_prompt.pdf}
        \caption{Llama-2-7B}
    \end{subfigure}
    \begin{subfigure}{.48\linewidth}
        \includegraphics[width=\linewidth]{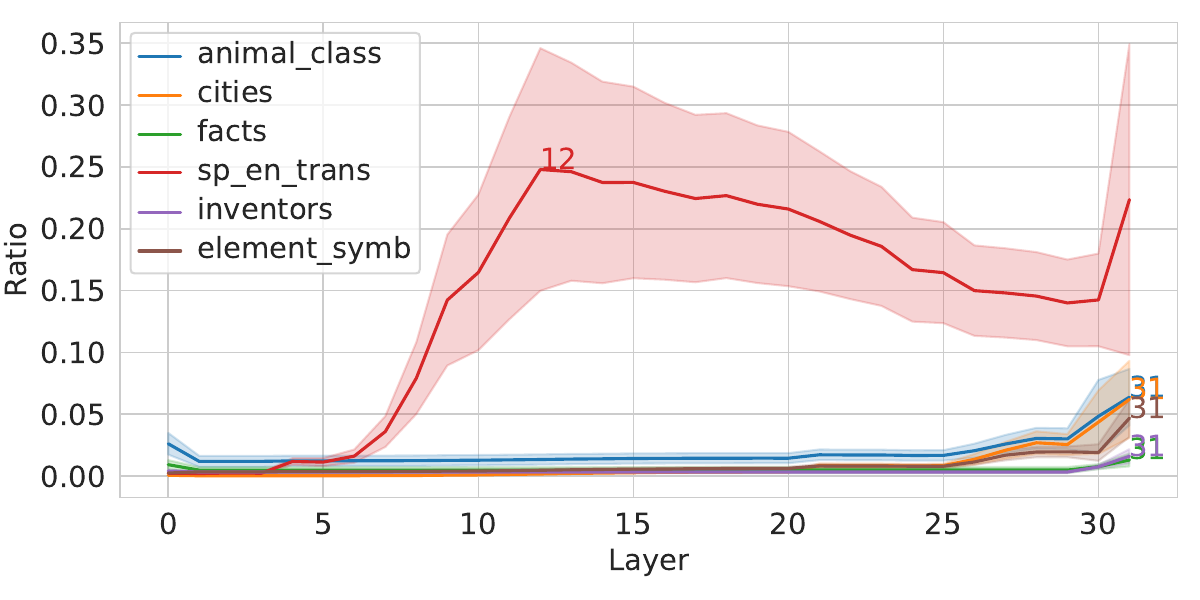}
        \caption{Llama-2-7B-Chat}
    \end{subfigure}
    
    \begin{subfigure}{.48\linewidth}
        \includegraphics[width=\linewidth]{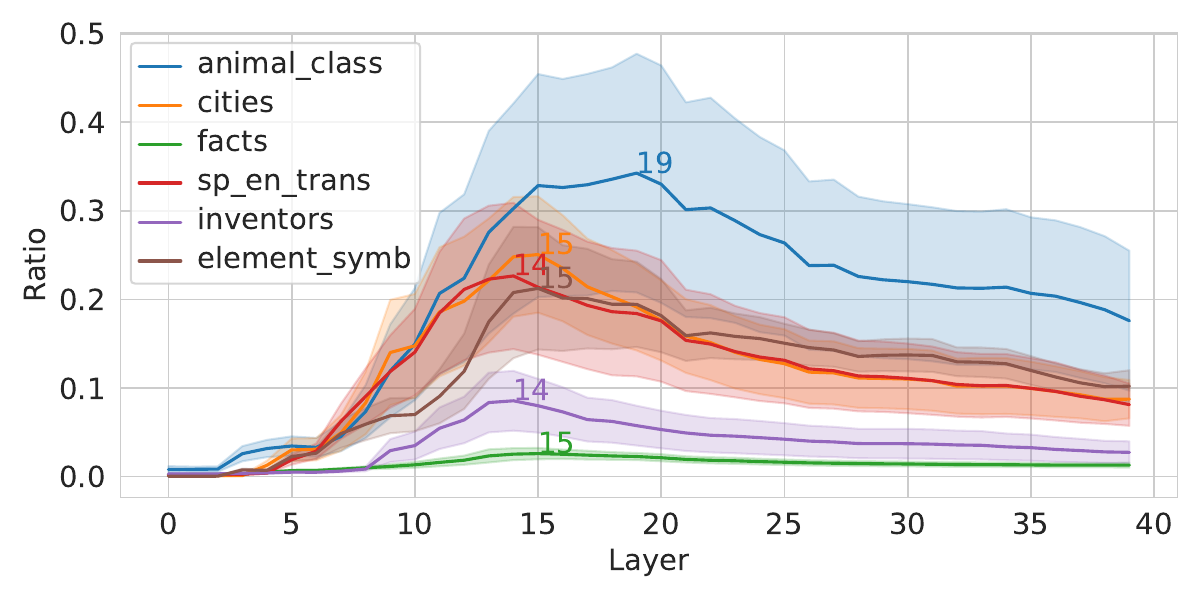}
        \caption{Llama-2-13B}
    \end{subfigure}
    \begin{subfigure}{.48\linewidth}
        \includegraphics[width=\linewidth]{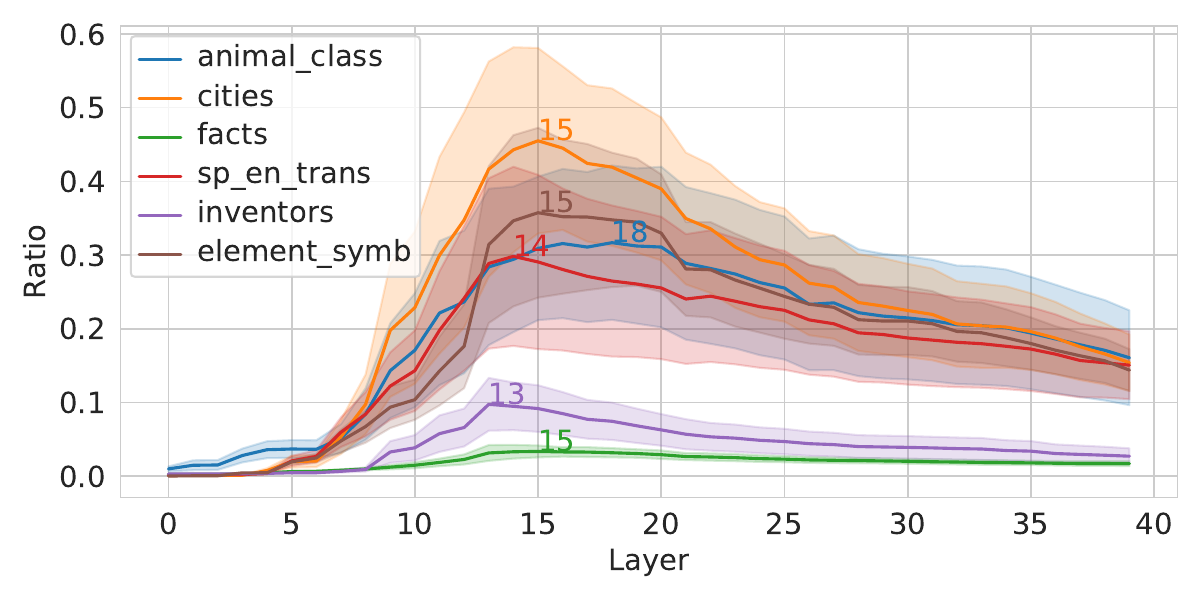}
        \caption{Llama-2-13B-Chat}
    \end{subfigure}
    
    \begin{subfigure}{.48\linewidth}
        \includegraphics[width=\linewidth]{figures/Meta-Llama-3.1-8B-hf_no_prompt.pdf}
        \caption{Llama-3.1-8B}
    \end{subfigure}
    \begin{subfigure}{.48\linewidth}
        \includegraphics[width=\linewidth]{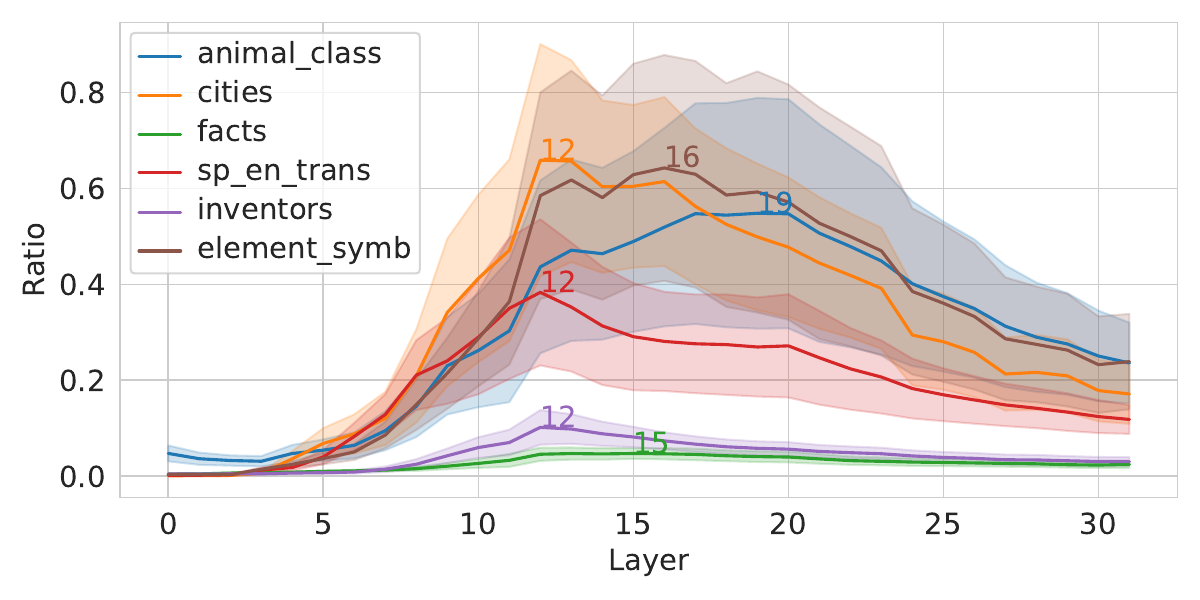}
        \caption{Llama-3.1-8B-Instruct}
    \end{subfigure}
    
    \begin{subfigure}{.48\linewidth}
        \includegraphics[width=\linewidth]{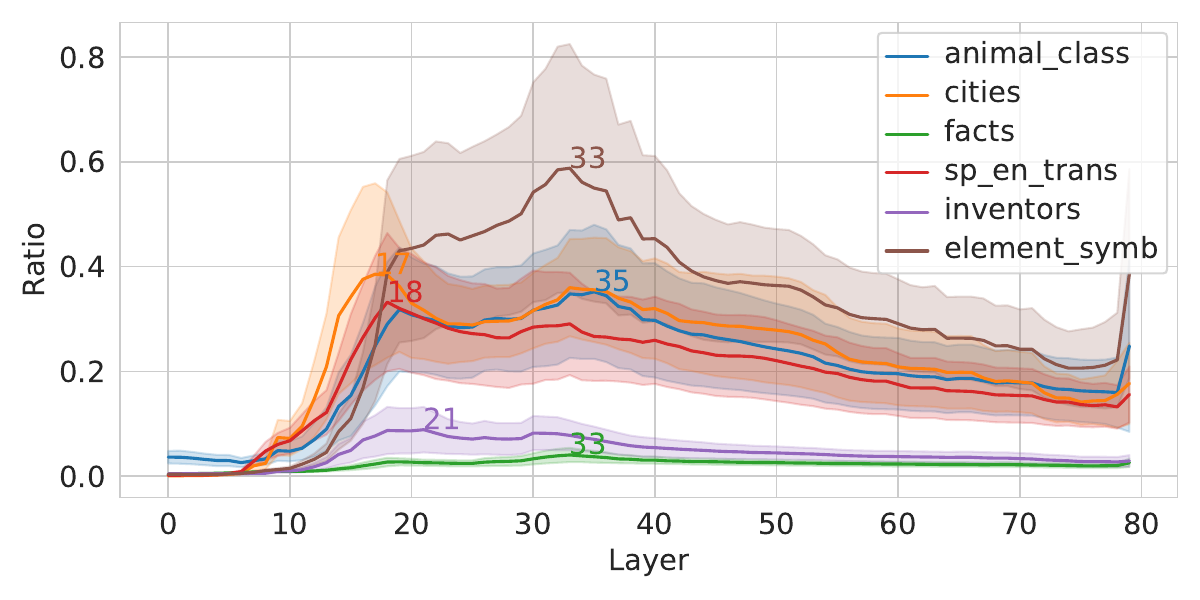}
        \caption{Llama-3.1-70B}
    \end{subfigure}
    \begin{subfigure}{.48\linewidth}
        \includegraphics[width=\linewidth]{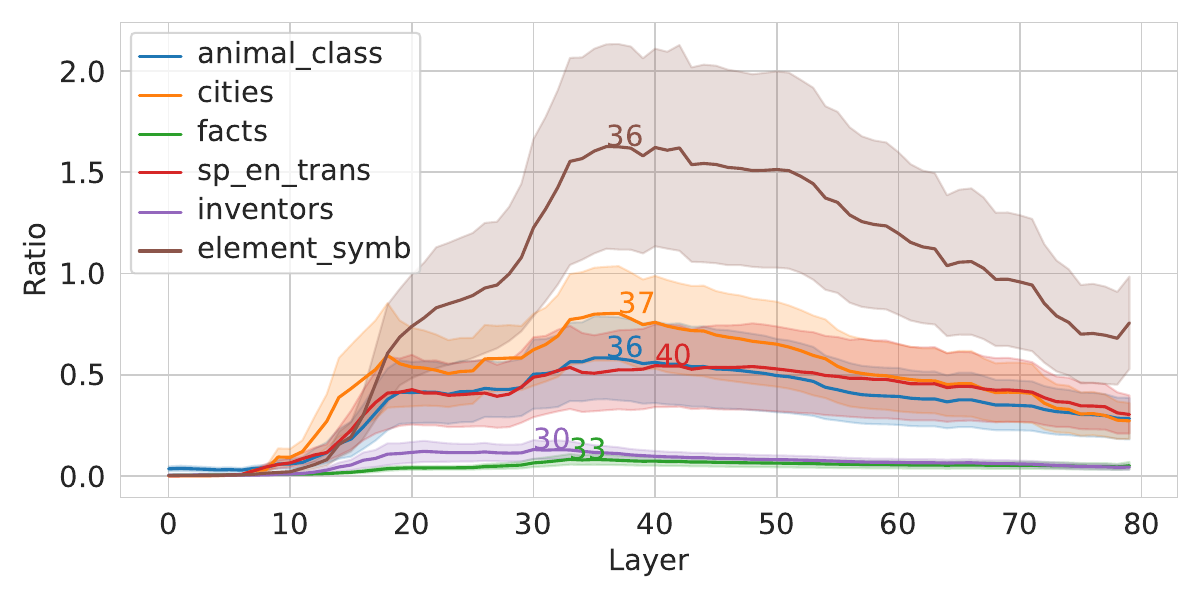}
        \caption{Llama-3.1-70B-Instruct}
    \end{subfigure}

    \begin{subfigure}{.48\linewidth}
        \includegraphics[width=\linewidth]{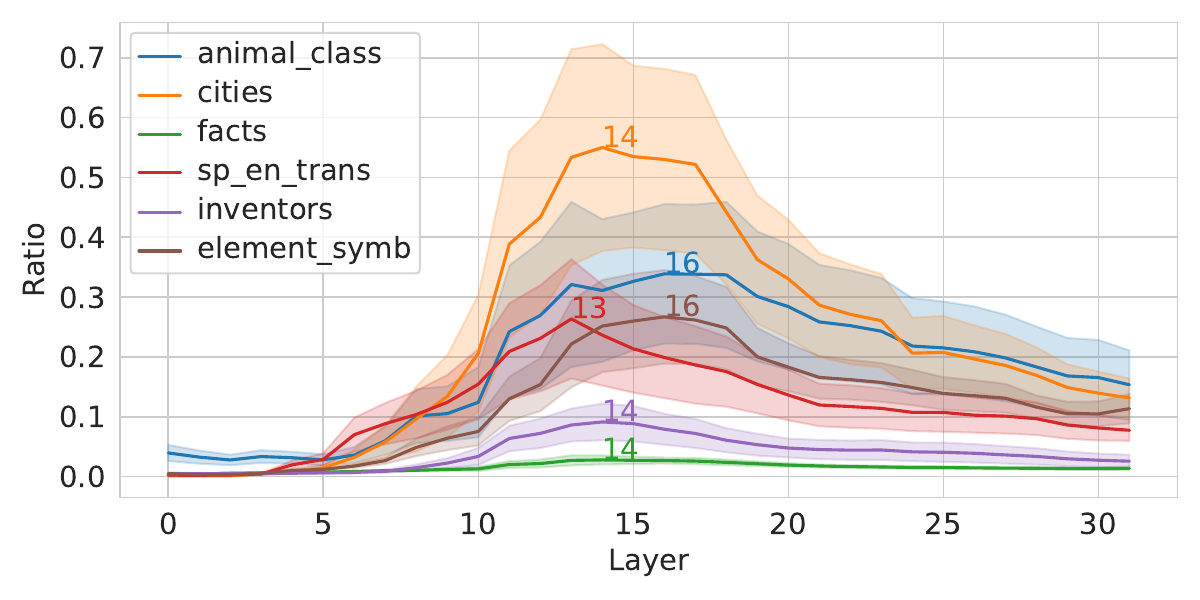}
        \caption{Mistral-7B-v0.1}
    \end{subfigure}
    \begin{subfigure}{.48\linewidth}
        \includegraphics[width=\linewidth]{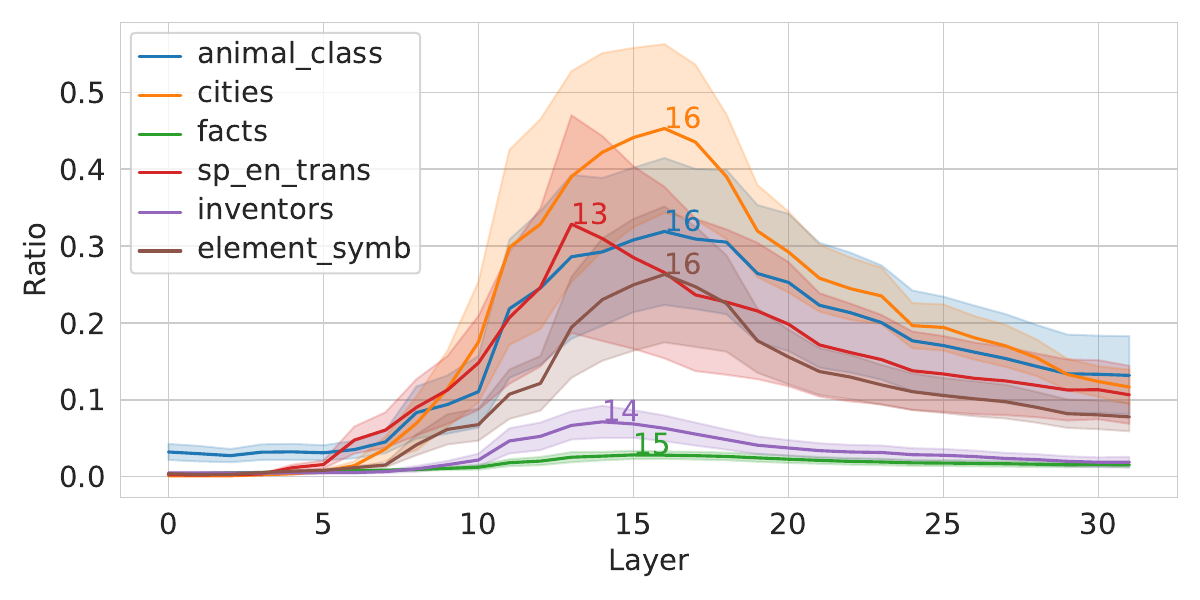}
        \caption{Mistral-7B-Instruct-v0.1}
    \end{subfigure}
    \caption{Plot of the ratio of between-class variance to within-class variance for a series of models. The shaded regions denote standard error.}
    \label{fig:layer_selection_full}
\end{figure*}

We summarize the plots in Figure \ref{fig:layer_selection_full}, which covers eight models: Llama-2-7B(-Chat), Llama-2-13B(-Chat), Llama-3.1-8B(-Instruct), Llama-3.1-70B(-Instruct), Mistral-7B(-Instruct)-v0.1. Their optimal layers are 12(13), 13(13), 12(13), 33(33), 13(13), respectively.

\subsection{Consistency of Truth Direction}
\begin{figure}
    \centering
    \includegraphics[width=1.\linewidth]{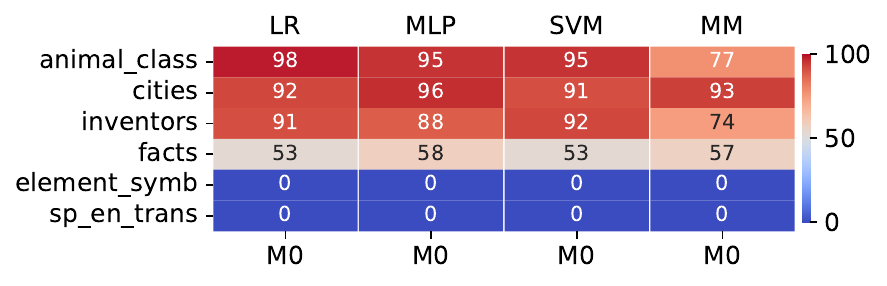}
    \caption{AUROC (in percentage) of probes trained on affirmative statements and tested on negative ones. AUROC exceeding 0.5 indicates generalization success. M0 refers to the Mistral-7B-v0.1 model.}
    \label{fig:neg_generalization_mistral_7b}
\end{figure}

We present results on Mistral-7B-v0.1 in Figure \ref{fig:neg_generalization_mistral_7b}. It is evident that the truthfulness probes generalize across negation on four topics, barely generalizing on \verb|facts| topic. Comparing these results with those of Figure \ref{fig:neg_generalization}, we notice that the performance of probes for Mistral-7B-v0.1 is comparable to that of probes for Llama-2-13B-Chat and Llama-3.1-8B. This aligns with the observation that the general capability of Mistral-7B-v0.1 lies between that of Llama-2-13B-Chat and Llama-3.1-8B.

\subsection{Logical Conjunction/Disjunction}
\begin{figure*}
    \centering
    \includegraphics[width=1.\linewidth]{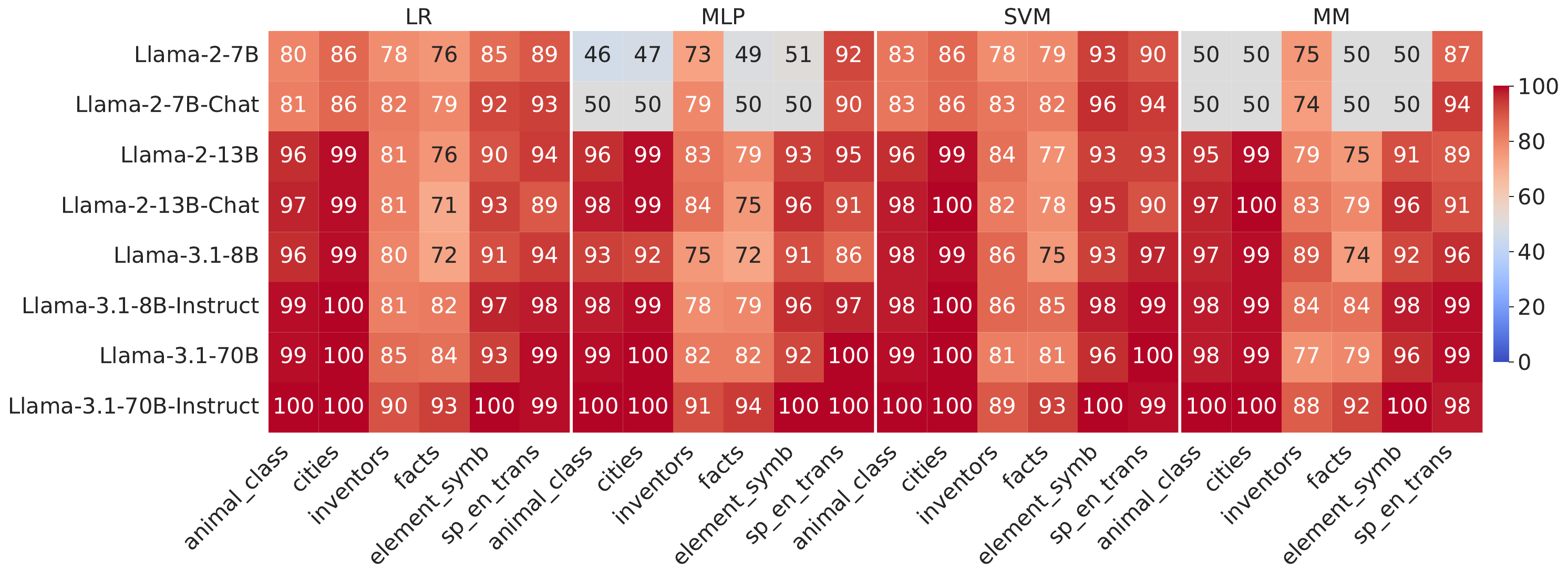}
    \caption{AUROC (in percentage) of probes trained on all the atomic factual statements and tested on logical conjunctions. $\text{AUROC}>0.5$ indicates the success of generalization.}
    \label{fig:conj_generalization_full}
\end{figure*}

\begin{figure*}
    \centering
    \includegraphics[width=1.\linewidth]{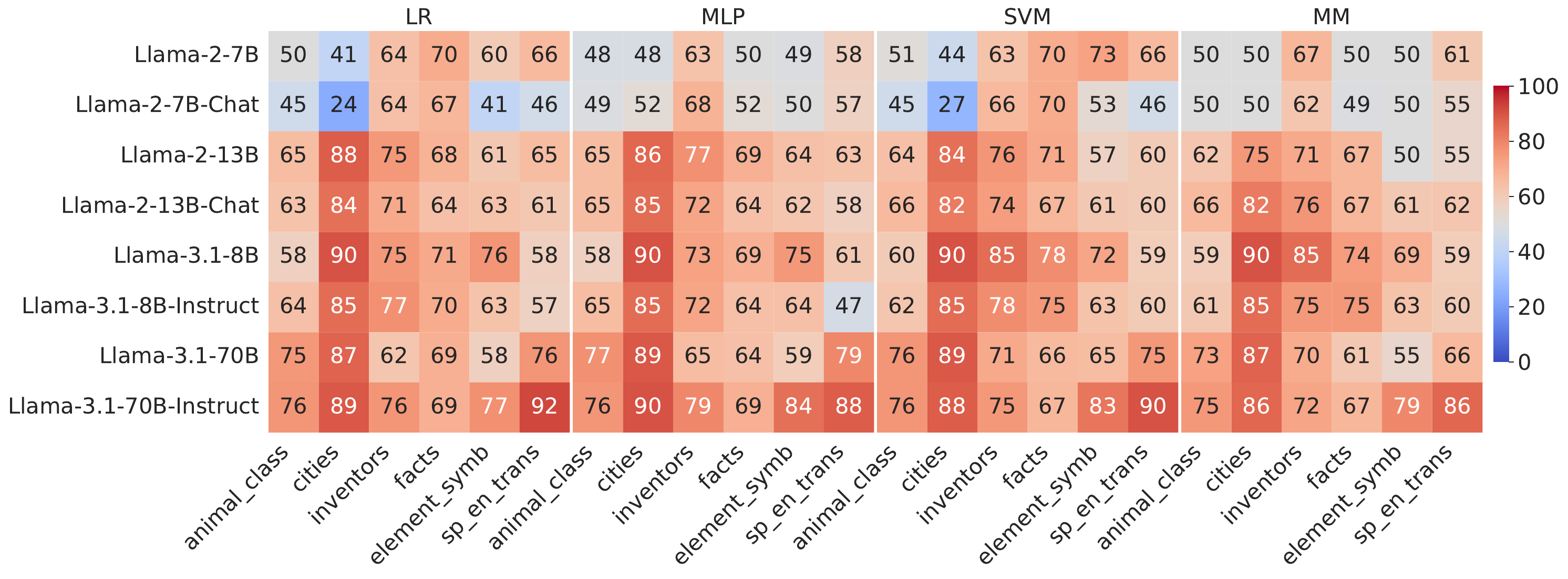}
    \caption{AUROC (in percentage) of probes trained on all the atomic factual statements and tested on logical disjunctions. $\text{AUROC}>0.5$ indicates the success of generalization.}
    \label{fig:disj_generalization_full}
\end{figure*}

Full results on logical conjunctions and disjunctions are shown in Figure \ref{fig:conj_generalization_full} and Figure \ref{fig:disj_generalization_full} respectively. A similar scaling trend could be observed as in Figure \ref{fig:neg_generalization}, where the classification accuracy of the probes is positively correlated with the target LLM's general capability.

\begin{figure}
    \begin{subfigure}{0.45\linewidth}
        \centering
        \includegraphics[width=\linewidth]{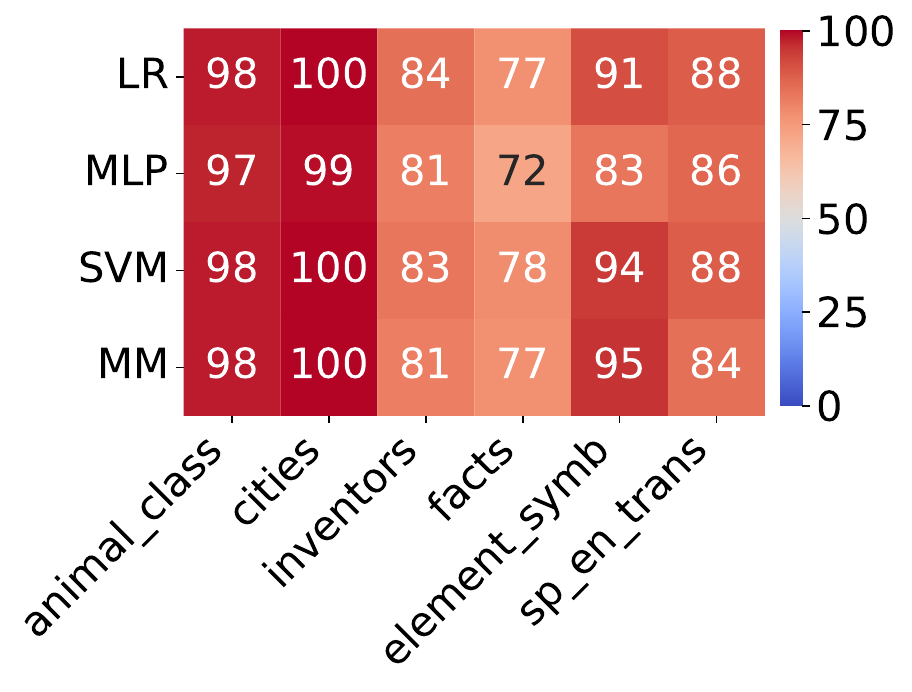}
        \caption{\centering Logical conjunctions.}
    \end{subfigure}
    \begin{subfigure}{0.45\linewidth}
        \centering
        \includegraphics[width=\linewidth]{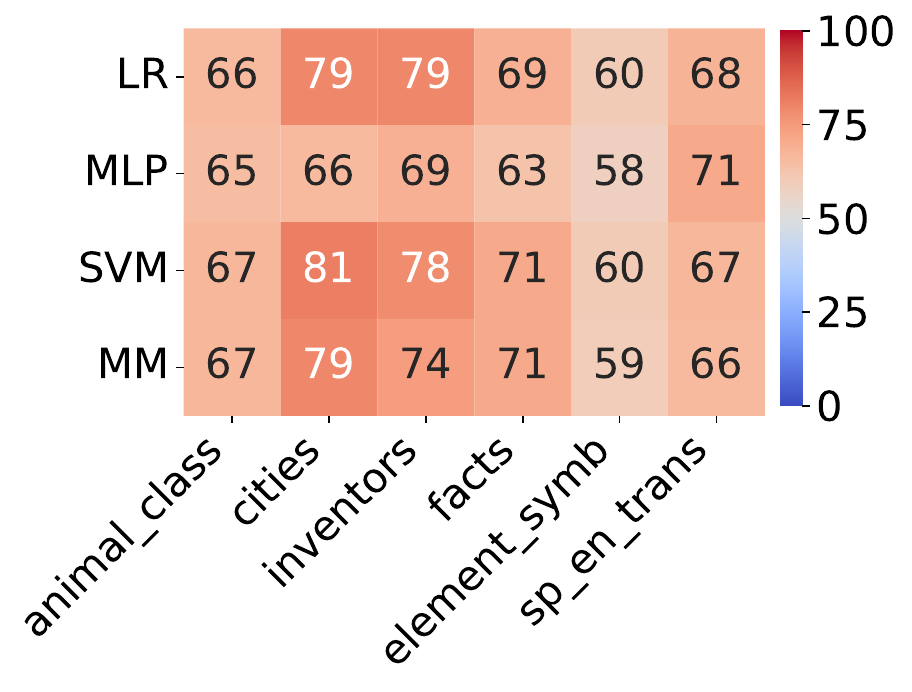}
        \caption{\centering Logical disjunctions.}
    \end{subfigure}
    \caption{AUROC (in percentage) of probes trained on atomic factual statements and tested on logical conjunctions/disjunctions for Mistral-7B-v0.1. $\text{AUROC}>0.5$ indicates the success of generalization.}
    \label{fig:generalization_to_conj_disj_mistral}
\end{figure}

The results for Mistral-7B-v0.1 is shown in Figure \ref{fig:generalization_to_conj_disj_mistral}. The truthfulness probes generalize from atomic factual statements to both logical conjunctions and disjunctions.

\subsection{Question Answering}
\subsubsection{Details on Experiment Setup}
\paragraph{MMLU.}
We arrange three setups for the QA task, and we demonstrate the prompt template for zero-shot setting using an actual example from the MMLU dataset. The few-shot prompts are trivially extended from the zero-shot prompt, with exemplars separated by two newlines (\verb|"\n\n"|). Few-shot exemplars are randomly selected from the development split.

\begin{lstlisting}[breaklines,basicstyle=\ttfamily\small]
Question: What was GDP per capita in the United States in 1850 when adjusting for inflation and PPP in 2011 prices?
Options:
A. About $300
B. About $3k
C. About $8k
D. About $15k
Answer: B
\end{lstlisting}

\paragraph{TriviaQA.}
For TriviaQA \citep{joshi2017triviaqa} we only use few-shot prompting -- 5-shot and 20-shot -- to ensure that the LLM always generates short-form answers. We use normalized answers in the exemplars.
\begin{lstlisting}[breaklines,basicstyle=\ttfamily\small]
Question: Where in England was Dame Judi Dench born?
Answer: york
\end{lstlisting}

\subsubsection{More results}
\begin{figure}[htbp]
    \centering
    \begin{subfigure}{1.\linewidth}
        \centering
        \includegraphics[width=1.\linewidth]{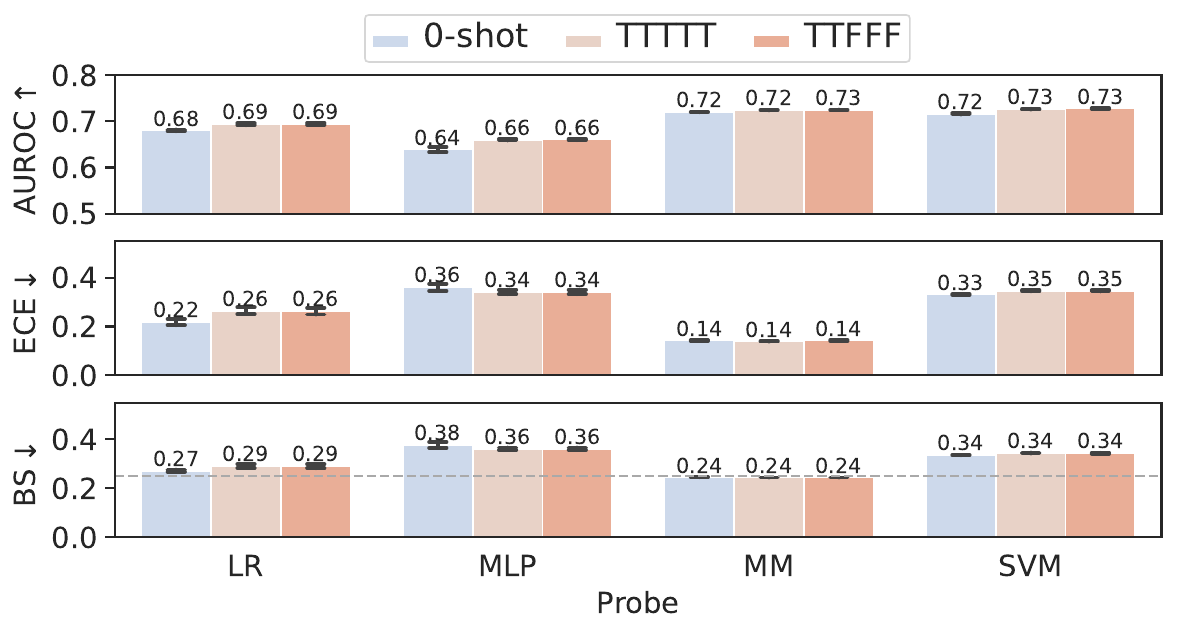}
        \caption{MMLU.}
    \end{subfigure}
    \begin{subfigure}{1.\linewidth}
        \centering
        \includegraphics[width=1.\linewidth]{figures/triviaqa.pdf}
        \caption{TriviaQA.}
    \end{subfigure}
    \caption{AUROC$\uparrow$/ECE$\downarrow$/BS$\downarrow$ of truthfulness probes for Mistral-7B-v0.1 on MMLU and TriviaQA. The dashed gray line corresponds to random results, and error bars denote standard error.}
    \label{fig:generalization_from_stmt_to_qa_mistral}
\end{figure}

The results for Mistral-7B-v0.1 is shown in Figure \ref{fig:generalization_from_stmt_to_qa_mistral}. The general behavior of truthfulness probes is similar to that in Figure \ref{fig:generalization_from_stmt_to_qa_mmlu}. The classification accuracy improves in response to few-shot prompting and improves when more in-context exemplars are provided in the prompt. Meanwhile, calibration only improves in the case of the SVM probe on TriviaQA dataset.

\subsection{Contextual Knowledge}
\subsubsection{Details on Experiment Setup}
\paragraph{SciQ.}
For the SciQ \citep{welbl2017crowdsourcing} benchmark, we arrange three setups, including ``zero-shot'', ``TTT'' and ``TTF''. We only demonstrate the zero-shot prompt as the few-shot prompts can be trivially extended from it. In the in-context setups, the exemplars are randomly selected from the training split.

\begin{lstlisting}[breaklines,basicstyle=\ttfamily\small]
Context: <context>
Question: Compounds that are capable of accepting electrons, such as o 2 or f2, are called what?
Options:
A. Oxygen
B. residues
C. antioxidants
D. oxidants
Answer: D
\end{lstlisting}

\paragraph{BoolQ.}
For the BoolQ \citep{clark-etal-2019-boolq} benchmark, we arrange four setups, including ``no options'', ``with options'', ``T'' and ``F''. We only demonstrate the prompt for ``with options''. In the in-context setups, the exemplars are randomly selected from the training split.

\begin{lstlisting}[breaklines,basicstyle=\ttfamily\small]
Passage: <passage>
Question: does ethanol take more energy make that produces?
Options:
- Yes
- No
Answer: No
\end{lstlisting}

\paragraph{XSum.}
For this task, we arrange four setups, including ``zero-shot'', ``T'', ``TT'' and ``TTT''. We only demonstrate the zero-shot prompt as the one-shot and few-shot prompts can be trivially extended from it. In the in-context setups, the exemplars are randomly selected from the training split of XSum \citep{Narayan2018DontGM} dataset, and they are all deemed correct. False examples come from the XSum Hallucination Annotations \citep{maynez-etal-2020-faithfulness} dataset with 500 examples, which is paired with examples from the test split of XSum. Furthermore, we filter for examples no longer than the LLM's context window. For Llama-3.1-8B with a context length of \num[group-separator={,}]{8192}, we obtain the final test set of 998 examples whose labels are balanced.

\begin{lstlisting}[breaklines,basicstyle=\ttfamily\small]
Summarize this document: <doc>
Summary: Rory McIlroy moved to within a shot of joint leaders Victor Dubuisson and Jaco van Zyl after the third round of the Turkish Airlines Open.
\end{lstlisting}

\subsubsection{More results}
\begin{figure*}
    \centering
    \includegraphics[width=1.0\linewidth]{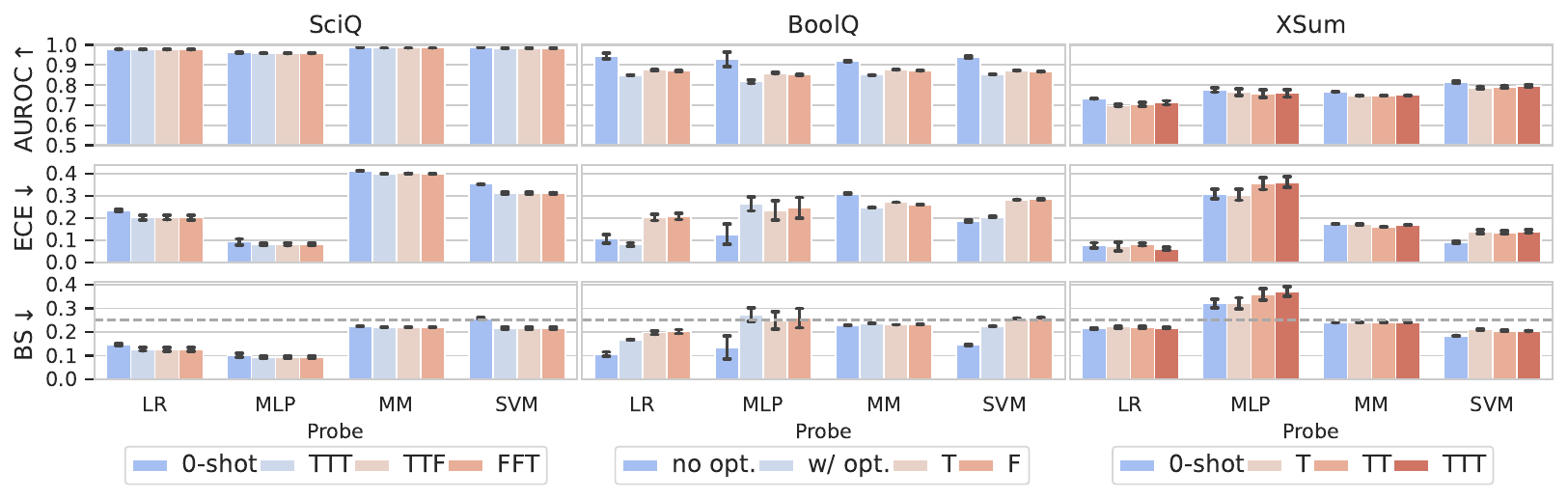}
    \caption{AUROC↑/ECE↓/BS↓ of truthfulness probes for Mistral-7B-v0.1 on tasks where grounding knowledge is
provided in the prompt. The dashed gray line corresponds to random results, and error bars denote standard error.}
    \label{fig:generalization_to_contextual_knowledge_mistral}
\end{figure*}

We present results for Mistral-7B-v0.1 in Figure \ref{fig:generalization_to_contextual_knowledge_mistral}. Accuracy improves as in-context exemplars are provided, but calibration only displays the same trend on SciQ dataset. Another abnormality could be observed for the BoolQ dataset from ``no options'' setting to ``with options'' setting, and for the XSum task from zero-shot to one-shot. In these cases, both accuracy and calibration worsens, which does not align with the results in Figure \ref{fig:generalization_from_paramatric_to_contextual}. We assume this is attributed to the weakness of the target model, where Mistral-7B-v0.1 finds it difficult to interpret answer options and in-context exemplars for the abstractive summarization task.

\section{License}
The implementation of the probes is based on the \verb|scikit-learn| \citep{scikit-learn} library, which is licensed under BSD 3-Clause License. The factual statements we use is curated by \citet{burger2024truth}, licensed under MIT License. The MMLU \citep{hendrycks2020measuring} dataset  is licensed under MIT License, the TriviaQA \citep{joshi2017triviaqa} dataset is licensed under Apache 2.0 License, the SciQ \citep{welbl2017crowdsourcing} dataset under Creative Commons Attribution-NonCommercial 3.0 Unported License, the BoolQ \citep{clark-etal-2019-boolq} under Creative Commons Share-Alike 3.0 License, the XSum \citep{Narayan2018DontGM} dataset under MIT License and the XSum Hallucination Annotations \citep{maynez-etal-2020-faithfulness} dataset under Creative Commons Attribution 4.0 International License. Llama-2 series of models are licensed under Llama 2 Community License Agreement, Llama-3 herd of models are licensed under Llama 3 Community License Agreement and Mistral models are licensed under MIT License.

\end{document}